\documentclass{article}

\usepackage[eandd,nonatbib,preprint]{neurips_2026}

\usepackage[numbers]{natbib}
\usepackage[utf8]{inputenc}
\usepackage[T1]{fontenc}
\usepackage{url}
\usepackage{booktabs}
\usepackage{amsfonts}
\usepackage{microtype}
\usepackage[x11names,table]{xcolor}
\usepackage{graphicx}
\usepackage{subcaption}
\usepackage{amsmath,amssymb}
\usepackage{bm}
\usepackage{algorithm}
\usepackage{algpseudocodex}
\usepackage{etoolbox}
\makeatletter\@ifpackageloaded{lineno}{\AtBeginEnvironment{algorithmic}{\nolinenumbers}}{}\makeatother
\usepackage{caption}
\usepackage{wrapfig}
\usepackage{multirow,multicol}
\usepackage{soul}
\usepackage{tabularx}
\IfFileExists{CJKutf8.sty}{\usepackage{CJKutf8}}{}
\usepackage{mathtools}
\usepackage{amsthm}
\usepackage{hyperref}
\hypersetup{
    colorlinks=true,
    linkcolor=blue,
    citecolor=blue,
    urlcolor=blue
}

\usepackage{titlesec}
\titlespacing*{\paragraph}{0pt}{0.5ex plus 0.2ex minus .2ex}{1em}
\titlespacing*{\section}{0pt}{1.4ex plus .5ex minus .2ex}{1ex plus .2ex}
\titlespacing*{\subsection}{0pt}{0.9ex plus .5ex minus .2ex}{.3ex plus .2ex}

\usepackage[capitalize,noabbrev]{cleveref}

\newcommand{\smol}{\fontsize{9}{9}\selectfont}
\newcommand{\shortminus}{\text{-}}
\newcommand{\graycomment}[1]{\textcolor{black!50}{\# #1}}
\newcommand{\hlc}[2][yellow]{{%
    \colorlet{foo}{#1}%
    \sethlcolor{foo}\hl{#2}}%
}

\theoremstyle{plain}

\theoremstyle{definition}

\theoremstyle{remark}

\usepackage[textsize=tiny]{todonotes}

\title{Searching the Internet for\\ Challenging Benchmarks at Scale}

\author{%
  Wenda Xu\thanks{Equal contribution.} \\
  Google\\
  \texttt{wendax@google.com} \\
  \And
  Vil\'em Zouhar$^*$ \\
  ETH Zurich\\
  \texttt{vzouhar@ethz.ch} \\
  \And
  Parker Riley \\
  Google\\
  \texttt{prkriley@google.com} \\
  \And
  Mara Finkelstein \\
  Google\\
  \texttt{marafin@google.com} \\
  \And
  Markus Freitag \\
  Google\\
  \texttt{freitag@google.com} \\
  \And
  Daniel Deutsch \\
  Google\\
  \texttt{dandeutsch@google.com} \\
}

\begin{document}

\maketitle

\begin{abstract}
Many static benchmarks are beginning to saturate: as models rapidly improve, they achieve near-perfect scores on fixed test sets, leaving little headroom to expose genuine model weaknesses---and even expert-curated challenge sets quickly saturate after hillclimbing.
We present a fully automatic framework that searches the Internet at scale to construct challenging benchmarks without human curation.
The key insight is to model the Internet as a vast space of topics and formalize the search as a multi-armed bandit problem, where each topic's difficulty is revealed only through expensive sample-and-evaluate queries.
Our $\epsilon$-greedy strategy identifies the most challenging topics while exploring only 6\% of the search space---a 100$\times$ cost reduction over exhaustive evaluation.
We validate on machine translation and knowledge question answering, confirming that discovered difficulty is robust across independent metrics (GEMBA-SQA and MetricX), languages, and models.
\end{abstract}

\begin{figure}[ht!]
    \centering
    \includegraphics[width=1\linewidth]{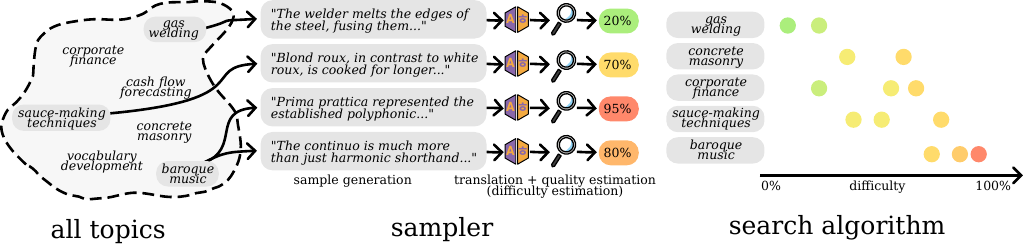}
    \caption{Illustration of our pipeline. Given a large set of all topics, the sampler can draw an example from a topic and estimate its difficulty. The goal of the search algorithm is to find the most difficult topic with as few samplings as possible.}
    \label{fig:pipeline}
    \vspace{-4mm}
\end{figure}

\section{Introduction}

Benchmark saturation is one of the most pressing challenges in modern LLM evaluation.
As models rapidly improve, static benchmarks lose their ability to expose genuine weaknesses: frontier models now achieve near-perfect scores in tasks such as machine translation \citep{kocmi-etal-2024-findings,kocmi-etal-2025-findings} and knowledge question answering \citep{goldman2025eclekticnovelchallengeset}, yet they clearly still fail on specialized or obscure content.
Even expert-curated challenge sets provide only temporary relief---they quickly saturate as models hillclimb, requiring repeated, costly manual effort to refresh.
This raises a fundamental question: \emph{can we automatically and continuously construct challenging benchmarks that expose genuine model headroom at scale?}
We answer yes, by treating the Internet as an effectively unbounded source of challenging content and searching it efficiently.

The Internet provides an effectively unbounded reservoir of complex and varied content, but manually curating it is non-reproducible, biased, and infeasible at scale.
Inspired by topic modeling \citep{blei2003}, we conceptualize the Internet as a structured topic space, where each text belongs to a particular topic and each topic contains a set of texts.
Our goal is to automatically search this space to identify the topics where current models still struggle---the regions beyond saturation where headroom remains.
The core challenge is that human intuition about topic difficulty does not reliably predict model failure modes: a topic's true difficulty can only be measured by observing where a model actually fails.
For example, a topic perceived as complex by humans---such as Baroque music theory---may be handled well by a model, while a seemingly mundane topic like concrete masonry can be a genuine failure point due to specialized terminology absent from training data (see \Cref{fig:pipeline}).
This misalignment between human-perceived and model-observed difficulty means we cannot rely on intuition or surface features to construct challenging benchmarks.
Determining difficulty requires an expensive sample-and-evaluate process: sampling a text from the topic, generating a model output, and estimating quality \citep{proietti-etal-2025-estimating}.
Applying this exhaustively across tens of thousands of topics would be computationally intractable.

We address this by formalizing the search as a multi-armed bandit (MAB) problem, where each topic is an ``arm'' whose difficulty is revealed through costly queries.
Our goal is to allocate a fixed budget of queries to identify the topics that yield the most consistently challenging examples---a task known as best arm identification \citep{AudibertBM2010}.
This formulation enables a principled exploration-exploitation tradeoff: the search strategically concentrates queries on the most promising topics, efficiently discovering where models fail without exhaustively evaluating every candidate.

We demonstrate this pipeline on two tasks where saturation is frequently reported: machine translation, where frontier models score near-perfectly on WMT and FLORES \citep{kocmi-etal-2024-findings,kocmi-etal-2025-findings}, and knowledge question answering, where they excel on established benchmarks yet still fail on specialized knowledge \citep{goldman2025eclekticnovelchallengeset}.
These tasks are also ideal because they are well-defined, with clear evaluation criteria and reliable automated quality estimators.
Indeed, tasks prone to saturation are typically those where the community has invested in well-validated metrics, so the availability of a quality estimator and the need for our framework naturally co-occur.
Validation further confirms that discovered difficulty is not an artifact of the proxy metric, but generalizes robustly across independent metrics, languages, and models.

\section{Methods}
\label{sec:methods}

We describe a general framework for automatic benchmark construction from Internet data, instantiated on two tasks: machine translation (MT) and knowledge question answering (KQA).
The framework has three components: (1) constructing a topic space that organizes the Internet into searchable units, (2) sampling texts from topics and estimating their difficulty, and (3) searching for the most difficult topics efficiently.

\subsection{Topic Space Construction}

The first step is to define the space of topics $T$ to search over.
We construct $T$ hierarchically by starting with broad seed categories---``science,'' ``business,'' ``law,'' ``education,'' and ``culture''---and recursively specializing each into finer-grained subtopics.
For instance, ``business'' yields ``finance,'' ``business innovation,'' and ``globalization''; ``finance'' in turn yields ``corporate finance,'' and so on.
At each level, a prompted LLM generates five subtopics, and the process recurses five levels deep, yielding $|T|=3.2$k leaf topics (see \Cref{sec:prompt_topic} for the topic generation prompt).
This construction is generally inexpensive and can be replaced by any predefined taxonomy or a list of topics or domains.

\subsection{Sampling and Difficulty Estimation}

Given a topic space $T$, the framework operates through a sample-and-evaluate loop: for each query, we draw a text from a topic and score how difficult that text is for current models.


\paragraph{Drawing samples from topics.}
Drawing a sample $x$ from a topic $t$ is achieved through grounded generation.
Specifically, we enable \href{https://ai.google.dev/gemini-api/docs/google-search}{Google search tool calling} when prompting the LLM to retrieve content relevant to topic $t$.
The LLM issues a Google Search query, retrieves relevant snippets, and extracts the most on-topic content from the returned results.
To ensure grounding in real web content, we request that the LLM pair each extracted item with a source URL for provenance.

\paragraph{Estimating difficulty.}
Once a sample $x$ with source text $s$ is drawn, we generate model outputs and score their quality.
Our framework requires a quality estimator that, given a model's output $o = m(s)$ for input $s$, assigns a difficulty score $d_x \in [0, 100]$, where higher values indicate greater model failure.
We average this score across a set of models $M$ to obtain the difficulty of input $s$: $d_s = \frac{1}{|M|} \cdot \sum_{m\in M} \mathrm{qe}(s, m(s))$.
The framework is agnostic to the choice of quality estimator; different tasks require different instantiations.
For \textbf{machine translation}, a topic $t$ is, e.g., ``1990s business news''; a sample $x$ is a source-language text; and $d_x$ is the difficulty score.
We use GEMBA \citep{kocmi-federmann-2023-large} based on Gemini-2.5-Pro, an LLM-as-a-judge metric that assigns a scalar quality score for an input--output pair $(s, o)$; difficulty is computed as 100 minus the quality score (see \Cref{sec:prompt_qe} for the quality estimation prompt).
This approximation via multiple model outputs is known as artificial crowd \citep{zouhar-etal-2025-select}.
For \textbf{knowledge question answering}, a topic $t$ is a knowledge domain, a sample $x$ is a factual question-answer pair generated from that domain, and $d_x$ is the model's error rate, averaged across multiple model configurations (see \Cref{sec:experiments}).

Together, drawing a sample and estimating its difficulty constitutes a single query with unit cost; the search algorithms described next decide which topics to query and how to allocate a fixed budget across them.

\subsection{Finding Difficult Topics as a Multi-Armed Bandit}

A topic $t$ is a distribution.
A sample from a topic $x \sim t$ is a piece of input text and has an associated difficulty score $d_x$.
Drawing a single sample $x$ from topic $t$, generating a model output for it, and estimating its difficulty has a cost of 1, and $t^\star$ denotes the set of samples that have been drawn from $t$.
The difficult topic search task is: given $T = \{ t \}_{j=1}^{|T|}$, find $\mathrm{top}{\shortminus}k_{t \in T}\,\,\, \mathbb{E}[d_x | x \sim t]$ at example budget $B$, so $\sum_{t\in T} |t^\star| \leq B$.
We evaluate any algorithm that selects some $\hat{T} \subseteq T, |\hat{T}|=k$ with budget $B$ by $\frac{1}{k} \cdot \sum_{\hat{t} \in \hat{T}} \mathbb{E}[d_x | x \sim \hat{t}]$ (i.e. selects the $k$ topics with the highest average difficulty), where higher is better.
The concrete instantiations of $t$, $x$, and $d_x$ for machine translation and knowledge QA are defined in the previous subsection.

\begin{table}

\begin{minipage}{0.48\linewidth}
\hrule
\vspace{1mm}
\begin{algorithmic}[1]
\Statex \hspace{-5mm} \textbf{Bandit}($T$: topics, $B$: budget, $k$):
\State \textbf{while} {$\sum_{t\in T} |t^\star| < B$}
    \Statex \quad \graycomment{Choose domain to sample from.}
    \State \quad $t \leftarrow \mathrm{ChooseToSample}(T)$
    \State \quad $x \sim t, t^\star \leftarrow t^\star \cup \{x \}$
\Statex \graycomment{Select most difficult.}
\State \Return $\mathrm{top}{\shortminus}k_{t \in T} \frac{\sum_{x \in t^\star} d_x}{|t^\star|}$
\end{algorithmic}
\hrule
\vspace{2mm}
\vspace{-7pt}
\captionof{algorithm}{General algorithm for non-structured search as a multi-armed bandit. The stopping criterion is given implicitly by reaching the budget. The selection criterion is simply the node with lowest observed maximum. The $\mathrm{ChooseToSample}$ functions are instantiated by \Cref{alg:brute,alg:greedy_bandit,alg:epsilon_greedy_bandit}.}
\label{alg:general_bandit}
\end{minipage}
\hfill
\begin{minipage}{0.48\linewidth}
\hrule
\vspace{1mm}
\begin{algorithmic}[1]
\Statex \hspace{-5mm} \textbf{BruteChoose}($T$: topics, $c$: cap):
\State \Return $\mathrm{Uniform}(\{t | t\in T, |t^\star|<c\})$
\end{algorithmic}
\hrule
\vspace{2mm}
\vspace{-7pt}
\captionof{algorithm}{Brute algorithm samples from a random topic limited by cap $c$.}
\label{alg:brute}
\vspace{10pt}

\hrule
\vspace{1mm}
\begin{algorithmic}[1]
\Statex \hspace{-5mm} \textbf{GreedyChoose}($T$: topics, $c$: cap):
\State \textbf{if} $\exists t \in T: |t^\star| = 0$
\State \quad \Return $\mathrm{Uniform}(\{ t | t \in T, |t^\star| = 0\})$
\State \textbf{else}
\State \quad \Return $\arg \max_{t \in T, |t^\star|<c} \frac{\sum_{x \in t^\star} d_x}{|t^\star|}$
\end{algorithmic}
\hrule
\vspace{2mm}
\vspace{-7pt}
\captionof{algorithm}{Greedy algorithm first samples from all topics and then exploits the most difficult one limited by cap $c$.}
\label{alg:greedy_bandit}
\end{minipage}
\end{table}

\begin{table}

\hrule
\vspace{1mm}
\begin{algorithmic}[1]
\Statex \hspace{-5mm} \textbf{EpsilonGreedyChoose}($T$: topics, $c$: cap, $\epsilon$: exploration):
\State \textbf{if} $\exists t \in T: |t^\star| = 0 \,\,\wedge\,\, \mathrm{Rand}() < \epsilon$
\State \quad \Return $\mathrm{Uniform}(\{ t | t \in T, |t^\star| = 0\})$
\hfill\graycomment{Select a yet non-explored topic.}
\State \textbf{else}
\State \quad \Return $\arg \max_{t \in T, |t^\star|<c} \frac{\sum_{x \in t^\star} d_x}{|t^\star|}$
\hfill\graycomment{Exploit the most promising topic.}
\end{algorithmic}
\hrule
\vspace{1mm}
\vspace{-7pt}
\captionof{algorithm}{Epsilon-Greedy algorithm stochastically switches between exploitation and exploration. The exploitation is limited by cap $c$.}
\label{alg:epsilon_greedy_bandit}

\end{table}

\subsection{Search Algorithms}

The general form of finding the best $t\in T$ is shown in \Cref{alg:general_bandit}.
Repeatedly, until we reach a budget, we select a topic to sample from (pull an arm), and at the end select the topic with the highest observed difficulty.
For our task, we use a series of increasingly complex selection methods:

\paragraph{Brute-force (\Cref{alg:brute}).}
The most basic approach samples from a random topic at each step.
All methods share a hyperparameter $c$ that caps the maximum number of samples from a single topic to prevent budget waste on a single ``honeypot'' arm.
This approach is uninformed and wastes budget on topics that do not look promising.

\paragraph{Greedy (\Cref{alg:greedy_bandit}).}
Selects the topic that currently has the highest observed difficulty (pure exploitation).
This requires all topics to be sampled at least once before exploitation can commence, which can be prohibitively expensive.

\paragraph{$\mathbb{\epsilon}$-Greedy (\Cref{alg:epsilon_greedy_bandit}).}
Stochastically switches between exploration (sampling never-sampled topics, with probability $\epsilon$) and exploitation (sampling the current best topic, with probability $1-\epsilon$).
This enables exploitation to begin before all topics have been scored, making it the most budget-efficient method in practice.

We also consider a batched version of these algorithms by replacing steps 2 and 3 in \Cref{alg:general_bandit} with choosing top-$b$ topics at once.
This helps avoid local minima through regularization.

\paragraph{Contextual Bandit.}
We also explore a contextual bandit variant that interpolates difficulty from neighboring topics using keyword similarity (Jaccard index), allowing the algorithm to skip topics whose neighbors are all easy.
Although promising, $\epsilon$-greedy matches or outperforms this variant in our experiments (\Cref{sec:other_algorithms}), suggesting limited local difficulty correlation; we leave richer context representations as future work.

\section{Experiments}
\label{sec:experiments}

We first compare the proposed search algorithms on both tasks, then analyze the discovered topics relative to existing benchmarks, validate robustness across scale, languages, models, and proxy signals, and report cost.
We use the framework on two tasks that exemplify benchmark saturation.

\textbf{Machine translation.}
We search for English source texts and evaluate translation into Czech, Chinese, German, and Ukrainian using Google Translate,\footnote{\href{https://translate.google.com/}{translate.google.com}, accessed 07-2025.} Gemini 2.5 Pro \citep{comanici2025gemini25pushingfrontier}, and Gemma 3 \citep{gemmateam2025gemma3technicalreport}, covering three language families and diverse model types (see \Cref{sec:prompt_sampling,sec:prompt_translation,sec:prompt_qe} for sampling, translation, and quality estimation prompts).
The topic set contains $\sim$3,000 topics, each with up to 25 samples.
Difficulty is measured by GEMBA-SQA score \citep{kocmi-federmann-2023-large} averaged across models and languages.

\textbf{Knowledge question answering.}
We generate 6,000 knowledge domains, each containing 25 subtopics.
For each topic, search retrieves Google snippets, and the model synthesizes question-answer pairs with definitive answers (see \Cref{sec:prompt_qa_sampling,sec:prompt_qa_answer,sec:prompt_qa_autorater} for prompts; Appendix~\ref{sec:knowledge_qa} provides full details).
After filtering context-dependent questions, we retain 5,160 domains ($\sim$130k subtopics).
Difficulty is estimated by averaging errors across three Gemini configurations: without thinking, with thinking, and with search enabled.

In both tasks the goal is to efficiently identify top-$k$ most difficult topics within a topic set $T$ of thousands of candidates.

\begin{figure}[t!]
    \centering
    \includegraphics[width=0.8\linewidth]{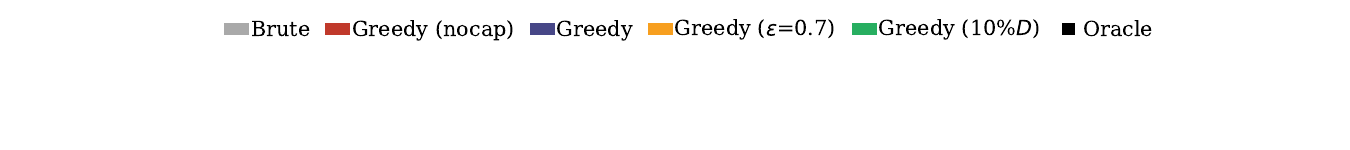}

    \vspace*{-10mm}
    \includegraphics[width=0.485\linewidth]{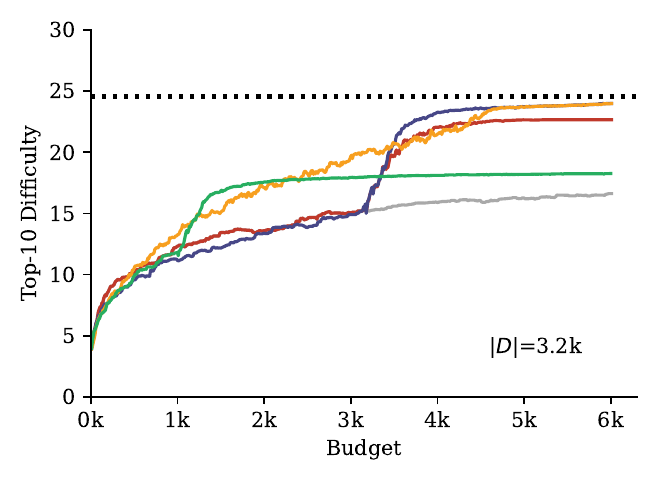}
    \includegraphics[width=0.485\linewidth]{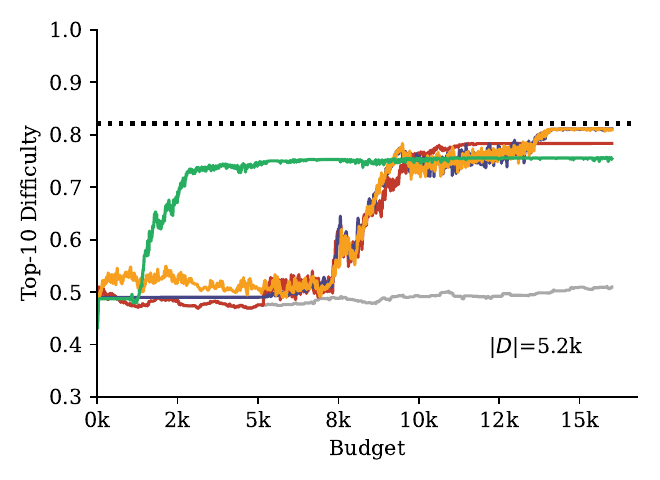}

    \vspace*{-4mm}
    \caption{Search algorithm comparison measured by top-10 difficulty on machine translation (left) and knowledge QA (right). All algorithms have the same budget; the cost of a single sampling is 1. Top-1 results are in Appendix~\ref{sec:top1_ablation}.}
    \label{fig:03-algorithms_main}
    \vspace{-2mm}
\end{figure}

\subsection{Search Algorithm Comparison}

We evaluate the search algorithms on both tasks by the difficulty of the final selected topic at each budget level (\Cref{fig:03-algorithms_main}).

\textbf{Machine translation.}
Brute search is near unusable, confirming that exploitation is essential.
Greedy works but requires sampling every topic at least once before exploitation can begin, which is prohibitively expensive.
$\epsilon$-greedy begins exploiting early and outperforms the greedy-on-subset baseline, which can miss difficult topics by chance.
The unconstrained greedy variant ($c=\infty$) gets stuck on a locally optimal topic, confirming the value of capping exploitation.
$\epsilon$-greedy reaches near-oracle difficulty ($\Delta<0.1$) by 4{,}500 steps---fewer than two samples per topic, corresponding to 6\% of the full evaluation cost.

\textbf{Knowledge QA.}
The trends closely mirror MT: bandit-based algorithms greatly outperform brute-force, and $\epsilon$-greedy recovers near-oracle difficulty by exploring only $\sim$10\% of the subtopic space ($\sim$13k out of 130k subtopics).

The takeaway is that $\epsilon$-greedy consistently finds near-oracle difficulty at a fraction of the cost across both tasks, confirming that the MAB formulation transfers without task-specific modification.
In \Cref{sec:other_algorithms} we discuss additional search algorithms.
Beyond top-$k$ performance, a critical question is whether the difficulty scores estimated during search faithfully reflect each topic's true difficulty.
\Cref{sec:estimated_vs_actual} shows that $\epsilon$-greedy's estimates converge to true per-topic difficulty far sooner than competing methods ($\Delta < 0.12$ by step 5{,}000), while brute-force never converges ($\Delta \approx 19$ even at step 6{,}000).

\begin{table}[ht!]
\centering
\small
\begin{minipage}[t]{0.5\linewidth}
\begin{tabular}{llrr}
\toprule
\multicolumn{2}{l}{\bf Existing Benchmarks} & \bf Difficulty$\bm{\uparrow}$ & \bf Words$\bm{\downarrow}$ \\
\midrule
\parbox[t]{4mm}{\multirow{4}{*}{\rotatebox[origin=c]{90}{\parbox{1.8cm}{\raggedleft WMT '24}}}}
& Social & 10.1 & 16 \\
& Literary & 8.3 & 38 \\
& Speech & 8.3 & 73 \\
& News & 6.4 & 54 \\[0.5em]
\parbox[t]{4mm}{\multirow{5}{*}{\rotatebox[origin=c]{90}{\parbox{1.8cm}{\raggedleft WMT '25}}}}
& Speech & 17.3 & 145 \\
& News & 14.3 & 95 \\
& Social & 11.8 & 98 \\
& Literary & 9.9 & 117 \\
& Dialogue & 5.7 & 179 \\[0.5em]
\parbox[t]{4mm}{\vspace{-2mm}\multirow{4}{*}{\rotatebox[origin=c]{90}{\parbox{2cm}{\raggedleft FLORES}}}}
& Disasters & 4.7 & 18 \\
& Travel & 4.6 & 21 \\
& Politics & 3.8 & 21 \\
& Sports & 3.6 & 18 \\
\bottomrule
\end{tabular}
\end{minipage}
\hfill
\begin{minipage}[t]{0.48\linewidth}
\begin{tabular}{lrr}
\toprule
\bf Our Topics (top-5) & \bf Diff.$\bm{\uparrow}$ & \bf Words$\bm{\downarrow}$ \\
\midrule
Prison vs. Jail & 39.5 & 29 \\
Leasehold Estates & 29.6 & 32 \\
Future Int.: Reversions & 29.5 & 34 \\
Removal Jurisdiction & 21.2 & 30 \\
Victim Impact Statements & 21.0 & 34 \\
\bottomrule
\end{tabular}
\end{minipage}

\vspace{3pt}
\caption{
Comparison of topics from existing benchmarks (left) and topics found by our $\epsilon$-greedy algorithm (right).
All discovered topics are more difficult than existing benchmark subsets despite having lower average word counts (difficulty scales with length).
The full table with example texts is in Appendix \Cref{tab:04-topic_overview_full}.
}
\label{tab:04-topic_overview}
\end{table}

\subsection{Discovered Challenge Topics Surpass Existing Benchmarks}


We compare against WMT 2024 \citep{kocmi-etal-2024-findings}, WMT 2025 \citep{kocmi-etal-2025-preliminary}, and FLORES-101 \citep{nllbteam2022} in \Cref{tab:04-topic_overview}.
Despite differences in granularity, sample length, and sample count, the oracle topics found by our search achieve comparable difficulty (error count) while being substantially shorter.
Appendix \Cref{tab:top_25-topic_overview} shows that our most challenging topic, ``Incarceration: Prison vs.\ Jail,'' achieves difficulty comparable to the hardest subsets across all three benchmarks, and all five top topics exceed most benchmark subsets.
Beyond the top-$k$, the full difficulty distribution shifts rightward compared to brute-force sampling (\Cref{fig:06-scores_across_nodes}), confirming that the algorithm improves difficulty identification across the entire distribution, not just at the extreme tail.
Our automatically discovered topics are harder than established benchmarks despite using shorter texts, demonstrating that the search finds genuinely challenging content rather than length-driven difficulty.
Having a more challenging test set is beneficial both for spotting model failures and for better benchmarking, which we discuss in \Cref{sec:ranking_efficiency}.

\begin{figure}[t!]
    \centering
    \includegraphics[width=0.9\linewidth]{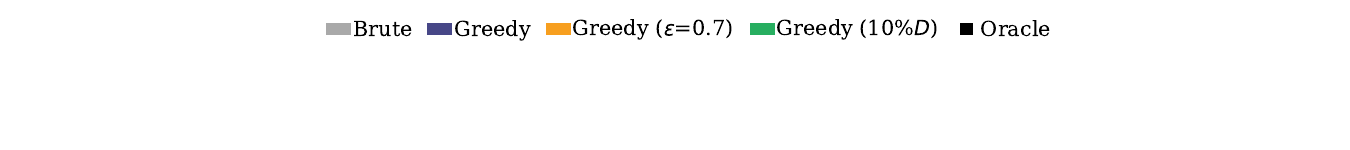}

    \vspace*{-11mm}
    \includegraphics[width=0.485\linewidth]{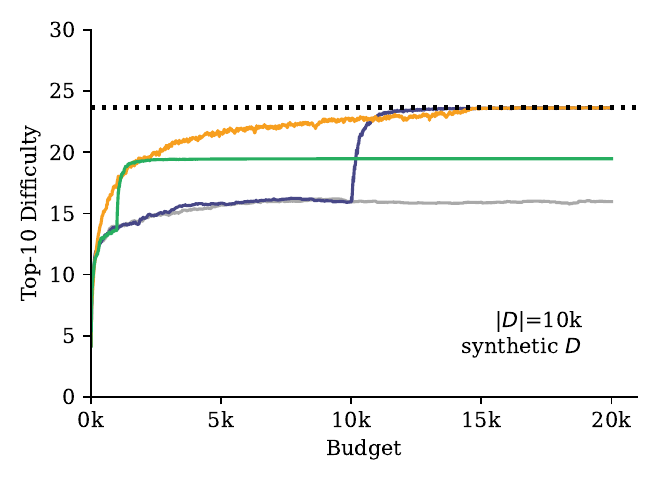}
    \includegraphics[width=0.485\linewidth]{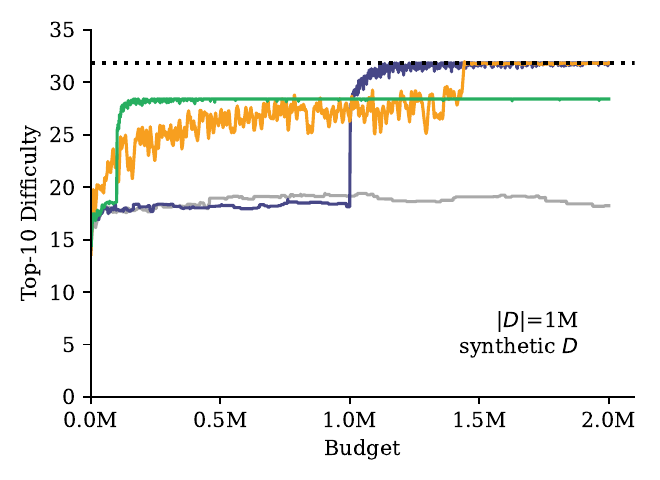}

    \vspace*{-3mm}
    \caption{Results for algorithms measured with top-10 difficulty on synthetically large $T$.
    All algorithms have the same budget and the cost of a single difficult estimation is 1.}
    \label{fig:03-algorithms_scale}
    \vspace{-4mm}
\end{figure}

\begin{figure}
    \centering
    \begin{minipage}{0.48\linewidth}
    \includegraphics[width=\linewidth]{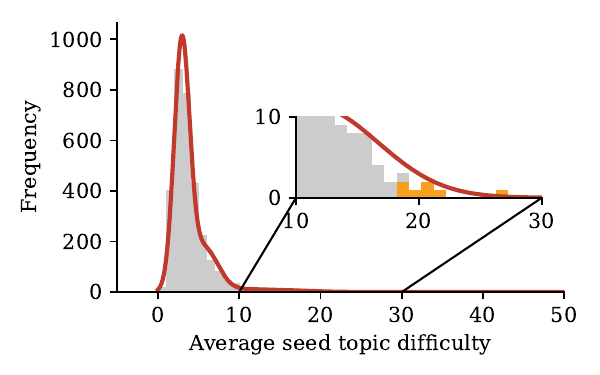}
    \caption{
    Distribution of topic difficulty (\hlc[gray!40]{gray}), best fit for Gaussian mixture model distribution with $3$ components (\hlc[red!60]{red}) used for synthetic scaling, and top-10 empirical oracle (\hlc[orange!80]{orange}).
    }
    \label{fig:06-scores_across_nodes}
    \end{minipage}
    \hfill
    \begin{minipage}{0.495\linewidth}
    
    \includegraphics[width=\linewidth]{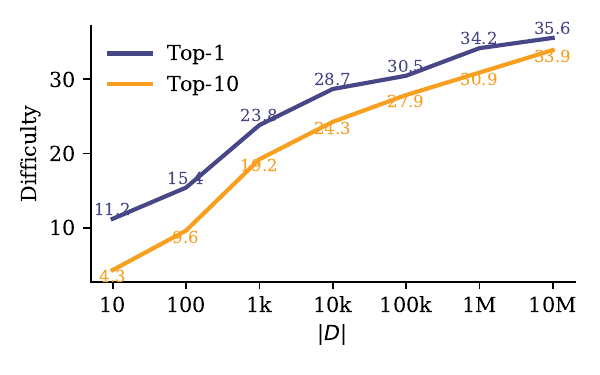}
    \vspace*{-8mm}
    \caption{Estimated oracle difficulty for topic sizes using a sample generative process as in \Cref{fig:06-scores_across_nodes}. The synthetic generation is more conservative than the real data (at $|T|=3.2$k the top-10 is 25 from \Cref{fig:03-algorithms_main}; see Appendix \Cref{fig:03-algorithms_top1} for top-1).
    }
    \label{fig:06-estimate_oracle_scale}
    \end{minipage}
\end{figure}

\subsection{Scaling to Larger Topic Spaces}

We verify scalability by synthesizing larger $T$: topic means $\mu_t$ are drawn from an empirically fitted Gaussian mixture (\Cref{fig:06-scores_across_nodes}), with $d_x \sim \mathcal{N}(\mu_t, \sigma^2)$.
\Cref{fig:03-algorithms_scale} shows that on synthetic sets of 10k and 1M topics, our algorithms reach near-oracle difficulty at $\sim$1.5 samples per topic ($\sim$6\%).
Difficulty grows approximately logarithmically with $|T|$ (\Cref{fig:06-estimate_oracle_scale}): a 10$\times$ increase yields roughly +5 difficulty, consistent with $\sqrt{\log |T|}$ extreme-value scaling \citep{leadbetter1983extremes}, confirming that the framework scales to arbitrarily large topic spaces with predictable gains.

\subsection{Monetary Cost of Benchmark Construction}
\label{sec:cost}

\begin{table}[t]
\centering
\small
\resizebox{0.8\linewidth}{!}{
\begin{tabular}{lllllll}
\toprule
\multicolumn{1}{c}{} & \multicolumn{4}{c}{\bf Cost \$} & \multicolumn{2}{c}{\bf Top-10 Difficulty} \\
\cmidrule(lr){2-5}
\cmidrule(lr){6-7}
\bf \# of Requests & \bf Search & \bf Translation & \bf QE & \bf Total & \bf Brute & \bf Greedy ($\epsilon${=}0.7) \hspace{-1mm} \\
\midrule
20k & \$87 &  \$2 & \$15 & \$104 & 16.3 & 19.7 \\
200k & \$867 & \$21 & \$152 & \$1040 & 17.9 & 25.3 \\
2M& \$8668 &  \$215 & \$1520 & \$10403 & 18.2 & 31.8 \\
\bottomrule
\end{tabular}
}
\vspace{2pt}
\caption{
At different monetary costs, $\epsilon$-greedy offers an advantage in identifying difficult topics over uninformed brute search.
}
\label{tab:real_cost}
\vspace{-5mm}
\end{table}

$\epsilon$-greedy consistently outperforms brute search at every budget level (\Cref{tab:real_cost}): for only \$104, it achieves top-10 difficulty of 19.7, which brute search cannot reach even at \$10{,}403, a 100$\times$ cost reduction.
Performance beyond our actual budget is estimated synthetically (\Cref{fig:06-scores_across_nodes,fig:06-estimate_oracle_scale}). A detailed token-level cost breakdown is provided in Appendix~\ref{sec:cost_full}.

While our pipeline uses Gemini-2.5-Pro as the difficulty estimator (\$1.25/1M input and \$10.00/1M output tokens), the framework is estimator-agnostic: the quality estimation component can be replaced with cheaper alternatives such as learned regression models (e.g., MetricX, COMET) or source-only difficulty predictors \citep{proietti-etal-2025-estimating}, reducing per-query cost while retaining the same MAB search framework.

\subsection{Robustness Across Languages and Models}
$\epsilon$-greedy consistently identifies near-optimal challenge samples across all four languages, including the high-resource English$\rightarrow$Chinese pair (Appendix \Cref{fig:07-algorithms_langs}).
Per-model results confirm robustness: $\epsilon$-greedy finds near-optimal challenge samples even against Gemini-2.5-Pro (Appendix \Cref{fig:07-algorithms_models_appendix}).
\Cref{tab:09-topic_similarity_languages} further shows that challenging topics transfer across related languages (e.g., Czech and Ukrainian), confirming that the discovered difficulty reflects intrinsically hard content rather than language- or model-specific artifacts.

\begin{table}[t!]
\fontsize{7.5pt}{9pt}\selectfont
\centering
\resizebox{\linewidth}{!}{\begin{tabular}{l@{\hspace{1mm}}l>{\raggedleft}p{9mm}>{\raggedleft}p{9mm}>{\raggedleft}p{9mm}>{\raggedleft\arraybackslash}p{9mm}}
\toprule
& & \multicolumn{4}{c}{Is $k$th of..} \\
& & \hspace*{-3mm} \bf Czech & \hspace*{-3mm} \bf Chinese & \hspace*{-3mm} \bf German & \hspace*{-3mm} \bf Ukrainian \\
\midrule
\parbox[t]{2mm}{\multirow{4}{*}{\rotatebox[origin=c]{90}{Top-10 of..}}}
& \bf Czech & \cellcolor{Yellow3!80}5.5 & \cellcolor{Yellow3!64}82.5 & \cellcolor{Yellow3!71}47.4 & \cellcolor{Yellow3!70}55.9 \\
& \bf Chinese & \cellcolor{Yellow3!42}190.3 & \cellcolor{Yellow3!80}5.5 & \cellcolor{Yellow3!17}313.7 & \cellcolor{Yellow3!61}100.0 \\
& \bf German & \cellcolor{Yellow3!53}136.3 & \cellcolor{Yellow3!10}345.4 & \cellcolor{Yellow3!80}5.5 & \cellcolor{Yellow3!59}108.8 \\
& \bf Ukrainian & \cellcolor{Yellow3!77}20.5 & \cellcolor{Yellow3!73}37.6 & \cellcolor{Yellow3!68}63.5 & \cellcolor{Yellow3!80}5.5 \\
\bottomrule
\end{tabular}
\begin{tabular}{ll>{\raggedleft}p{9mm}>{\raggedleft}p{9mm}>{\raggedleft\arraybackslash}p{9mm}}
\toprule
& & \multicolumn{3}{c}{Is k-th of..} \\
& & \bf Gemini & \bf Gemma & \bf G.Trans. \\
\midrule
\parbox[t]{2mm}{\multirow{3}{*}{\rotatebox[origin=c]{90}{Top-10..}}}
& \bf Gemini & \cellcolor{Yellow3!80}5.5 & \cellcolor{Yellow3!39}30.8 & \cellcolor{Yellow3!72}10.3 \\
& \bf Gemma & \cellcolor{Yellow3!10}48.4 & \cellcolor{Yellow3!80}5.5 & \cellcolor{Yellow3!12}47.3 \\
& \bf G.Trans. & \cellcolor{Yellow3!62}16.8 & \cellcolor{Yellow3!55}20.9 & \cellcolor{Yellow3!80}5.5 \\
\bottomrule
\vspace{0mm}
\end{tabular}}
\vspace{3pt}
\caption{Cross-language and -model topic difficulty analysis. Difficult topics are largely difficult across languages and models.
} 
\label{tab:09-topic_similarity_languages}
\vspace{-5mm}
\end{table}

\subsection{Cross-Metric Validation and Proxy Over-Optimization}
\label{sec:cross_metric}

We examine two related questions: (1) does our Gemini-based GEMBA-SQA difficulty estimator merely find topics that are hard for LLM-as-judge without reflecting genuine translation difficulty, and (2) does optimizing a proxy signal lead to over-optimization, and how can it be mitigated?

\paragraph{GEMBA-SQA-guided search finds genuinely harder data.}
A key concern is that the GEMBA-SQA difficulty estimator---which itself uses a Gemini model---might exploit Gemini-specific biases rather than surface true translation difficulty.
To rule this out, we validate using MetricX~\citep{juraska-etal-2023-metricx}, a state-of-the-art learned quality estimation metric that is fully independent of Gemini.
As shown in \Cref{tab:cross_metric_guide}, GEMBA-SQA-guided search also improves MetricX difficulty over the first 6{,}000 steps (MetricX score increases from 8.03 to 12.7), confirming that the discovered topics are genuinely harder under an independent metric.
This validates that the Gemini-based estimator is not hacking its own score: both signals agree that the search is finding progressively more challenging content.

\paragraph{Proxy signal hacking and mitigation.}
Despite this, using a single metric as the sole guide carries a risk of over-optimization.
As shown in \Cref{tab:cross_metric_guide}, using MetricX alone as the guide quickly maximizes MetricX difficulty (17.3 $\to$ 20.1) but simultaneously \emph{degrades} GEMBA-SQA scores (6.47 $\to$ 5.57)---a clear sign of metric-specific artifact exploitation.
This demonstrates that optimizing a single proxy can steer the search toward topics that expose weaknesses of that particular metric rather than genuine model failures.
We identify two effective mitigation strategies:
\textbf{(1) Ensemble architecturally diverse metrics.} Combining both signals by averaging GEMBA-SQA and MetricX scores yields the best trade-off across both metrics, confirming that ensembling metrics from fundamentally different model families is a robust defense against over-optimization.
\textbf{(2) Early stopping guided by a held-out metric.} One can use GEMBA-SQA as the primary guide but monitor MetricX as an independent validation signal and stop when MetricX no longer improves---analogous to early stopping in model training.
Both strategies require no additional sampling cost and generalize to any proxy-guided search framework.

\begin{table}[t]
\centering
\small
\begin{tabular}{r cc cc cc}
\toprule
 & \multicolumn{2}{c}{\textbf{MetricX Guide}} & \multicolumn{2}{c}{\textbf{GEMBA-SQA Guide}} & \multicolumn{2}{c}{\textbf{MetricX+GEMBA-SQA}} \\
\cmidrule(lr){2-3} \cmidrule(lr){4-5} \cmidrule(lr){6-7}
\textbf{Steps} & MetricX & GEMBA-SQA & MetricX & GEMBA-SQA & MetricX & GEMBA-SQA \\
\midrule
2000  & 17.3 & 6.47  & 11.9 & 16.6 & 14.1 & 12.2 \\
4000  & 19.0 & 5.98  & 12.4 & 20.8 & 14.6 & 15.2 \\
6000  & 19.8 & 5.86  & 12.7 & 23.2 & 15.3 & 18.0 \\
10000 & 19.9 & 5.57  & 12.7 & 23.3 & 15.3 & 18.0 \\
20000 & 20.1 & 5.74  & 12.7 & 23.6 & 15.3 & 18.0 \\
\bottomrule
\end{tabular}
\vspace{3pt}
\caption{Cross-metric validation under three guidance settings. MetricX Guide optimizes MetricX quickly but degrades GEMBA-SQA; GEMBA-SQA Guide steadily improves GEMBA-SQA but saturates MetricX; the combined MetricX+GEMBA-SQA guidance yields the best trade-off. Before search begins, MetricX and GEMBA-SQA difficulty scores are $3.56$ and $8.03$, respectively.}
\label{tab:cross_metric_guide}
\vspace{-5mm}
\end{table}

\subsection{Reproducibility with Open-Source Models}
\label{sec:open_source}

To demonstrate that the framework is not dependent on a frontier model, we replicate the pipeline using Gemma-27B-it~\citep{gemmateam2025gemma3technicalreport}, an open-source model, for all content generation steps: topic proposal, domain tree construction (depth~5, branching factor~5, yielding 3{,}200 topics), test sentence extraction, and difficulty estimation, using the same prompts as for the Gemini-based pipeline (see \Cref{sec:prompt_topic,sec:prompt_sampling,sec:prompt_translation,sec:prompt_qe}).
Then we use Gemini-2.5-Pro to evaluate top-10 domains that discovered by Gemma model at each budget level.

\begin{table}[t]
\centering
\small
\begin{tabular}{lcc}
\toprule
\textbf{Method} & \textbf{Step 6{,}000} & \textbf{Step 10{,}000} \\
\midrule
Brute force & 11.16 & 12.30 \\
Greedy, batched, $D$=10\% & 16.82 & 17.15 \\
Greedy, no cap & 18.25 & 18.38 \\
$\epsilon$-greedy, batched, $\epsilon$=0.7 & 20.83 & 22.13 \\
\bottomrule
\end{tabular}
\vspace{3pt}
\caption{Difficulty scores when using Gemma-27B-it for content generation. The relative ordering mirrors the Gemini-based results, confirming that the framework generalizes to open-source models.}
\label{tab:gemma_results}
\vspace{-5mm}
\end{table}

\Cref{tab:gemma_results} reports top-10 difficulty under this setup.
$\epsilon$-greedy achieves 20.83 at step 6{,}000 and 22.13 at step 10{,}000, compared to an oracle difficulty of 30.2.
The relative ordering and performance gaps between algorithms are consistent with the Gemini-based results (\Cref{fig:03-algorithms_main}), confirming that the MAB framework transfers effectively to open-source models, with the remaining gap attributable to content generation and evaluation capabilities.
This demonstrates that the entire pipeline can be reproduced without proprietary models, making the framework accessible to the broader research community.
Combined with \Cref{tab:real_cost}, this offers a practical path toward reproducible and cost-effective benchmark construction: ideally we can use open source models for all steps and perform this benchmark construction in-house.

\begin{table}[t]
    \centering
    \small
    \renewcommand{\arraystretch}{0.95}
    \resizebox{0.75\linewidth}{!}{
    \begin{tabular}{lr c lr c lr}
    \toprule
    \multicolumn{2}{c}{\textbf{Error Severity}} & \phantom{abc} & \multicolumn{2}{c}{\textbf{Error Category}} & \phantom{abc} & \multicolumn{2}{c}{\textbf{Error Type}} \\
    \cmidrule{1-2} \cmidrule{4-5} \cmidrule{7-8}
    Major & \cellcolor{Firebrick3!80} 70.7\% && Terminology & \cellcolor{Firebrick3!58} 47.7\% && Inappropriate for Context & \cellcolor{Firebrick3!58} 47.6\% \\
    Minor & \cellcolor{Firebrick3!33} 21.3\% && Accuracy & \cellcolor{Firebrick3!53} 42.3\% && Mistranslation & \cellcolor{Firebrick3!47} 35.9\% \\
    Critical & \cellcolor{Firebrick3!21} 8.0\% && Style & \cellcolor{Firebrick3!18} 5.1\% && Omission & \cellcolor{Firebrick3!19} 5.6\% \\
    & && Fluency & \cellcolor{Firebrick3!18} 4.9\% && Awkward & \cellcolor{Firebrick3!19} 5.5\% \\
    & && & && Untranslated & \cellcolor{Firebrick3!17} 3.6\% \\
    & && & && Grammar & \cellcolor{Firebrick3!15} 1.8\% \\
    \bottomrule
    \end{tabular}
    }
    \vspace{3pt}
    
     \caption{AutoMQM \citep{fernandes-etal-2023-devil} error breakdown for 250 challenge examples discovered by $\epsilon$-greedy (averaged across four language directions and three models). The benchmark predominantly surfaces terminology and accuracy errors---the genuine remaining weaknesses of SOTA MT systems---rather than fluency or style issues that modern LLMs have largely solved.}
    \label{fig:error_type_analysis}
    \vspace{-7mm}
    
\end{table}

\section{What Does Our Benchmark Expose?}
\label{sec:error_analysis}

A natural question is whether the challenge examples surfaced by our pipeline reveal genuine model weaknesses or merely reflect biases in the search process.
We use AutoMQM \citep{fernandes-etal-2023-devil} (prompt in \Cref{sec:prompt_automqm}) to systematically categorize the errors in our collected test set across three translation models and four language directions.\footnote{While GEMBA-SQA assigns a scalar quality score used to guide the difficulty search, AutoMQM provides a complementary fine-grained analysis by marking error locations with error category and severity labels based on Multidimensional Quality Metrics (\citealp{freitag-etal-2021-experts}). Because AutoMQM is independent of the search process, we use it here to analyze the error distribution of the discovered benchmark data.}

As shown in \Cref{fig:error_type_analysis}, 90\% of errors fall into two categories: \emph{terminology} (47.7\%) and \emph{accuracy} (42.3\%), manifesting as contextually inappropriate word choices and outright mistranslations.
Fluency and style errors account for only $\sim$10\%.
This distribution is not a limitation of our pipeline but a faithful reflection of where SOTA MT systems actually fail: modern LLM-based translators have largely solved fluency, and the remaining hard frontier is domain-specific terminology and factual accuracy.
This is independently corroborated by \citet{kocmi-etal-2025-findings}, who report that SOTA systems still struggle with linguistic complexity and domain-specific terminology.

Crucially, the error profile is a property of each model, not of the input.
Different models produce markedly different error distributions on the same test set: Gemma-3-27b produces more fluency errors (14.2\% vs.\ 8.7\%), while Gemini-2.5-Pro shifts toward accuracy errors (53.7\% vs.\ 37.6\%).
Appendix~\ref{sec:error_profile_analysis} provides a detailed analysis of these contrasting error profiles, showing that Gemini-2.5-Pro's strength in fluency comes at the cost of higher accuracy errors, while Gemma-3-27b exhibits the reverse pattern---confirming that each model has distinct weaknesses that our benchmark exposes.
If our inputs lacked structural complexity, all models would exhibit near-identical profiles; the observed variation confirms that the challenge examples are rich enough to trigger distinct failure modes in different systems.
The framework is also estimator-agnostic: replacing the quality estimator with a fluency-specific metric would shift the search toward fluency errors instead.
Appendix \Cref{tab:example_ende,tab:example_encs,tab:example_enzh} presents concrete examples of these errors.

\section{Related Work}
\label{sec:related_work}

\paragraph{Difficult examples.}
The example difficulty is tied to evaluating the example-level quality of model outputs.
In machine translation, this is commonly done with reference-free automated metrics, such as SEScore \citep{xu-etal-2022-errors, xu-etal-2023-sescore2}, InstructScore \citep{xu-etal-2023-instructscore}, COMET \citep{rei-etal-2020-comet} or MetricX \citep{juraska-etal-2023-metricx}, which provide a score corresponding to the output quality.
This score can be used as a proxy for difficulty.
\citet{proietti-etal-2025-estimating} train a model that predicts the expected model performance based on just the source, which can be used for difficulty estimation.
Knowing the difficulty of an example for models has many uses, ranging from curriculum learning \citep{jia-etal-2025-semantic} to more efficient evaluation \citep{zhan-etal-2021-difficulty}.
Approaches for searching for difficult examples are often limited to some apriori knowledge of difficulty, which guides the selection of syntactically complex texts or texts with rare words \citep{chen-etal-2023-multifaceted}.
A different approach generates, not searches, for difficult-to-translate texts \citep{pombal2025zeroshot,lu2025automated,zouhar-etal-2025-generating} or otherwise adversarial examples \citep{zhang-etal-2021-crafting,sadrizadeh-etal-2024-classification}.

\paragraph{Bandits.}
Our task setup in \Cref{sec:methods} corresponds to the Best Arm Identification problem \citep{AudibertBM2010}.
Many works also make use of some features of the arms, such as their similarities, to inform the choices \citep{li2010contextual}.
In machine translation, \citet{cheng-etal-2025-bayesian,zouhar-etal-2025-early} use a generalization of the multi-armed bandit to improve machine translation and quality estimation efficiency.
Also in machine translation, \citet{kumar-etal-2019-reinforcement,kreutzer-etal-2021-bandits-dont} use bandit formulation for training data selection and curriculum learning.

\paragraph{Active and curriculum learning.}
Our framework is fundamentally distinct from both active learning and curriculum learning.
Active learning \citep{settles2009active} selects examples from a \emph{fixed} pool to label; our method searches an \emph{unbounded} space---the Internet---to \emph{discover} new examples.
Curriculum learning \citep{bengio2009curriculum} orders known training examples by difficulty; our method identifies \emph{which topics are difficult} in the first place.
The difficult examples we discover could serve as inputs to either approach, making them complementary (see Appendix~\ref{sec:related_work_full} for a detailed discussion).

\section{Conclusion}

Static benchmarks are saturating, and future progress demands evaluation that can keep pace with rapidly improving models.
We have shown that the Internet, when searched efficiently, provides a rich source of challenging content that lies beyond the reach of existing benchmarks.
By formalizing this search as a multi-armed bandit problem, our fully automatic pipeline identifies the most difficult topics while exploring only 6\% of the search space---a 100$\times$ reduction in cost compared to exhaustive search.
The automatically constructed benchmarks surpass the difficulty of established test sets like WMT and FLORES for machine translation, and we demonstrate identical gains on knowledge question answering, confirming the task-agnostic nature of the framework.
Cross-metric validation with MetricX confirms that discovered difficulty reflects intrinsic model weaknesses rather than proxy artifacts.
When proxy over-optimization is a concern, ensembling metrics or applying early stopping on a held-out validation signal are both effective and cost-free mitigations.
The framework is robust across languages and models, and the full pipeline can be reproduced with open-source models, making it accessible to the broader research community.
Our work establishes a path from static, manually curated benchmarks to dynamic, automatically constructed ones that evolve alongside model capabilities.
Future work should extend this approach to additional tasks (e.g., mathematical reasoning, code generation, safety) or integrate it into an active learning setup where discovered challenging examples are used for targeted training rather than solely for evaluation.
\bibliographystyle{unsrtnat}
\bibliography{anthology,main}

\newpage
\appendix

\section{Ablation: Top-1 Topic Performance}
\label{sec:top1_ablation}

\begin{figure}[t!]
    \centering
    \includegraphics[width=0.8\linewidth]{img/03-algorithms_legend.pdf}

    \vspace*{-10mm}
    \includegraphics[width=0.485\linewidth]{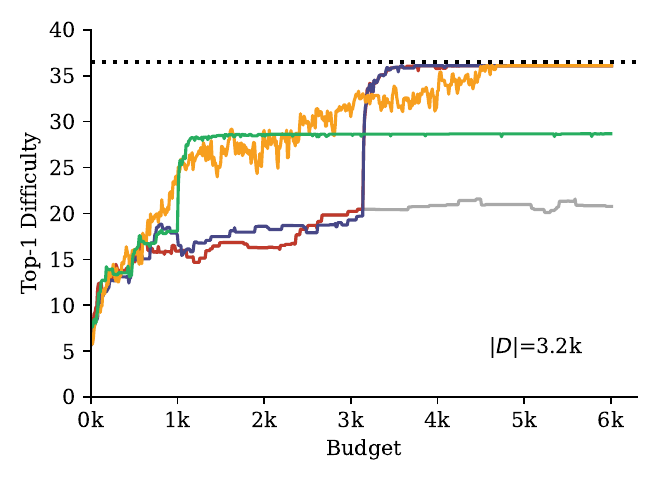}
    \includegraphics[width=0.485\linewidth]{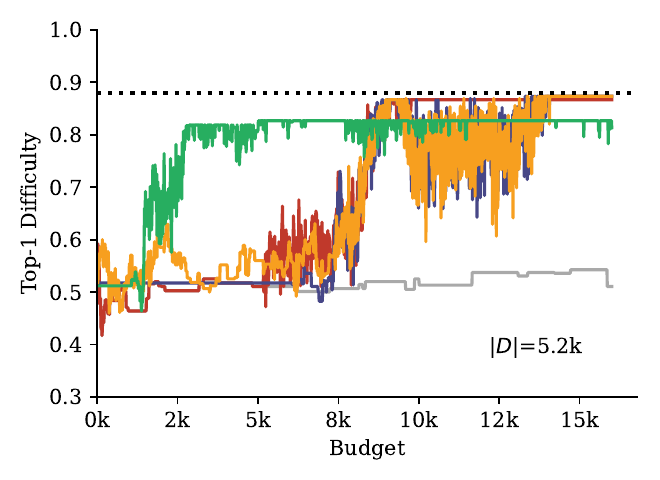}

    \vspace*{-5mm}
    \caption{Search algorithm comparison measured by top-1 difficulty on machine translation (left) and knowledge QA (right). All algorithms have the same budget; the cost of a single sampling is 1.}
    \label{fig:03-algorithms_top1}
\end{figure}

In the main paper (\Cref{fig:03-algorithms_main}), we report search algorithm performance using top-10 difficulty.
Here we complement that analysis with top-1 results on both tasks (\Cref{fig:03-algorithms_top1}), which measure how well each algorithm identifies the single most difficult topic.

The trends are consistent with the top-10 results: $\epsilon$-greedy dominates across both tasks, reaching near-oracle difficulty with a fraction of the budget.
Notably, top-1 performance is noisier than top-10 because it depends on a single topic rather than an average over ten, making the advantage of $\epsilon$-greedy's balanced exploration-exploitation even more pronounced.
On machine translation, $\epsilon$-greedy converges to within $\Delta < 1$ of the oracle by step 5{,}000.
On knowledge QA, the same pattern holds, with $\epsilon$-greedy recovering oracle-level difficulty by exploring only $\sim$10\% of the subtopic space.
These results confirm that the findings from the main paper (\Cref{fig:03-algorithms_main}) are robust to the choice of $k$ and generalize across both evaluation tasks.

\section{Robustness Across Languages and Models}
\label{sec:robustness_appendix}

In this section, we provide detailed per-language and per-model breakdowns of the search algorithm performance, complementing the aggregate results in the main paper (\Cref{sec:experiments}).

\begin{figure}[t!]
    \centering
    \includegraphics[width=0.8\linewidth]{img/03-algorithms_scale10k_legend.pdf}
    \vspace{-10mm}
    
    \includegraphics[width=0.8\linewidth]{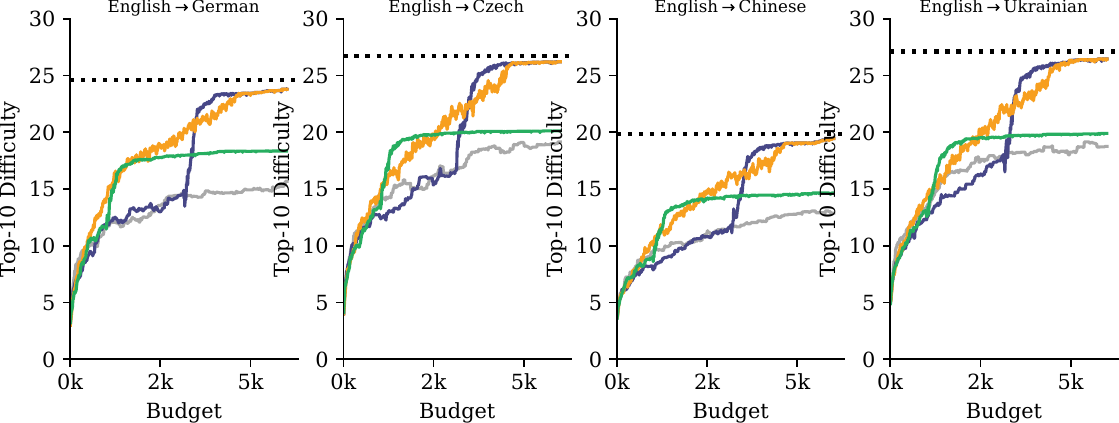}
    \caption{Our search algorithms perform consistently across all four language directions. Per-model results are in \Cref{fig:07-algorithms_models_appendix}.}
    \label{fig:07-algorithms_langs}
\end{figure}

\paragraph{Per-language results.}
\Cref{fig:07-algorithms_langs} shows algorithm performance separately for English$\rightarrow$Czech, English$\rightarrow$Chinese, English$\rightarrow$German, and English$\rightarrow$Ukrainian.
$\epsilon$-greedy consistently identifies near-optimal challenge topics across all four language directions, including the high-resource English$\rightarrow$Chinese pair where baseline performance is already strong.
The convergence rate is similar across languages, confirming that the search framework does not favor particular language families.

\begin{figure}[t!]
    \centering
    \includegraphics[width=0.8\linewidth]{img/03-algorithms_scale10k_legend.pdf}
    \vspace{-10mm}
    
    \includegraphics[width=0.8\linewidth]{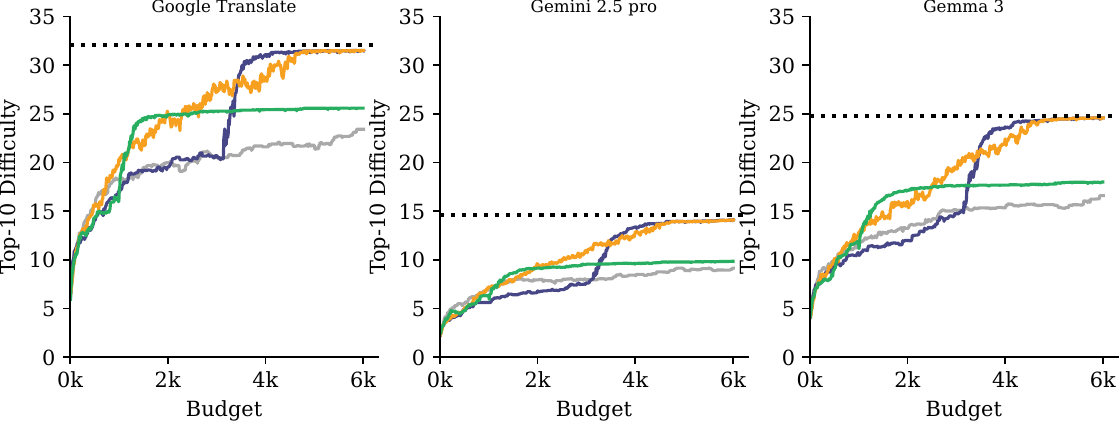}
    \caption{Our search algorithms consistently identify challenging source texts across Gemini-2.5-Pro, Gemma3-27B, and Google Translate.}
    \label{fig:07-algorithms_models_appendix}
\end{figure}

\paragraph{Per-model results.}
\Cref{fig:07-algorithms_models_appendix} breaks down performance by translation model: Gemini-2.5-Pro, Gemma3-27B, and Google Translate.
The relative algorithm ordering is preserved across all three models, with $\epsilon$-greedy dominating in each case.
Notably, even against Gemini-2.5-Pro---a frontier model that is also used as our quality estimator---the search successfully identifies genuinely challenging topics, ruling out the concern that the framework merely finds topics where the estimator model struggles.

\section{Relationship to Active Learning and Curriculum Learning}
\label{sec:related_work_full}

Our framework is fundamentally distinct from both active learning and curriculum learning.
Active learning \citep{settles2009active} selects examples from a \emph{fixed} pool to label; our method searches an \emph{unbounded} space---the Internet---to \emph{discover} new examples.
Curriculum learning \citep{bengio2009curriculum} orders known training examples by difficulty; our method identifies \emph{which topics are difficult} in the first place.
The difficult examples we discover could serve as inputs to either approach, making them complementary.
Our MAB formulation corresponds to Best Arm Identification \citep{AudibertBM2010}; we emphasize that the contribution is the \emph{problem formulation} rather than a novel bandit algorithm---even simple $\epsilon$-greedy yields 100$\times$ savings over brute-force in this previously un-formalized setting.

\section{Knowledge Question Answering}
\label{sec:knowledge_qa}
To demonstrate the generalization capabilities of our proposed pipeline, we extend our framework to knowledge question answering, following the approach of ECLeKTic \citep{goldman2025eclekticnovelchallengeset}. Analogous to our machine translation pipeline, we use the prompt in \Cref{sec:prompts} to generate $6,000$ domains, each containing 25 distinct topics. For each topic, we employ a search agent to formulate queries and retrieve the most relevant Google snippets. Conditioned on this context, we instruct the model---using the prompt in \Cref{sec:prompts}---to synthesize factual question-response pairs. Inspired by ECLeKTic, we target factual questions with definitive answers. Crucially, while the model relies on the retrieved snippets to generate the pairs, we explicitly instruct it to formulate questions that can be answered without access to the context. Finally, to ensure data consistency, we filter out any domain containing at least one topic where the questions remain context-dependent. Ultimately, we obtain a dataset of $5,160$ domains, each containing $25$ subtopics. 

Since the question-response pairs are synthesized directly from source snippets, we treat them as ground truth and employ reference-based evaluation for the model-generated answers (see \Cref{sec:prompts} for specific prompts). Autorater scale range is between $0$ to $1$. Therefore, difficult estimation is $1$ - quality estimation. We include three model to answer factual questions: 1) Gemini without thinking. We set the thinking budget to $0$ when sampling answers. 2) Gemini with thinking. We use the default thinking budget when sampling answers. 3) Gemini with search turned on. The final quality estimation is calculated by averaging the performance across the responses from these three models.

The knowledge QA trends closely mirror the machine translation results (\Cref{fig:03-algorithms_main}), as discussed in the main paper. The top-1 performance for knowledge QA is shown in \Cref{fig:03-algorithms_top1}, and the top-10 performance is in \Cref{fig:03-algorithms_main} (right panel). Bandit-based algorithms greatly outperform the brute-force approach, further demonstrating the generalizability of our framework. Greedy and $\epsilon$-Greedy surpass other bandit methods, demonstrating the value of exploitation. Although Greedy (10\%) excels at rapidly locating sub-optimal domains in early steps, $\epsilon$-Greedy achieves higher performance with a larger budget. Most importantly, we recover the oracle difficult samples by exploring only 10\% of the total search space ($13k$ out of $130k$ subtopics)---equating to fewer than $3$ samples per domain.

\section{Estimated vs.\ Actual Difficulty}
\label{sec:estimated_vs_actual}

A natural concern about bandit-based search is whether the difficulty scores estimated \emph{during} search accurately reflect the true difficulty that would be measured with a much larger, independent sample per domain.
To address this, we compare the estimated difficulty (i.e., the running proxy score used to guide search) against the actual difficulty (i.e., the independently measured per-domain difficulty score) for all four algorithms at six budget checkpoints spanning 1{,}000 to 6{,}000 steps.
We quantify the gap as $\Delta = |\text{Estimated} - \text{Actual}|$; smaller values indicate more faithful estimation.

\Cref{tab:estimated_vs_actual} summarizes the results.
Three insights emerge:

\paragraph{$\epsilon$-Greedy converges fastest.}
$\epsilon$-greedy ($\epsilon$=0.7) achieves $\Delta = 3.65$ as early as step 1{,}000 and $\Delta < 0.12$ by step 5{,}000.
In contrast, Greedy (no cap)---which always exploits by sampling from the currently hardest topics---incurs a large early overestimation ($\Delta = 25.00$ at step 3{,}000) before resolving it by step 4{,}000 ($\Delta = 0.50$).
By balancing exploration and exploitation, $\epsilon$-greedy avoids being misled by high-variance arms that initially appear difficult but have moderate true difficulty.
Both methods converge to near-zero $\Delta$ by step 6{,}000, but $\epsilon$-greedy's estimates are reliable much sooner.

\paragraph{Brute-force never converges.}
Brute-force maintains $\Delta \approx 19$--$25$ throughout the entire search ($\Delta = 19.19$ even at step 6{,}000), confirming that without guided exploration, estimated difficulty remains wildly inflated relative to actual difficulty.
With a budget of ${\sim}6{,}000$ samples spread uniformly across 35k+ topics, each topic receives fewer than one sample on average; the resulting top-$k$ selection is dominated by topics that received high scores by chance from very few observations---exactly the high-variance, moderate-difficulty topics that bandit algorithms are designed to avoid.

\paragraph{Top-$D$ achieves accurate estimates but is capped by its restricted search space.}
Top-$D$ ($D$=10\%) attains near-zero $\Delta$ as early as step 2{,}000 ($\Delta = 0.09$), demonstrating good estimation accuracy.
However, because it restricts search to only the top 10\% of topics, its actual difficulty plateaus at 18.26---significantly below $\epsilon$-greedy's 23.98 and the oracle's 24.56.
Despite reliable estimation within its restricted subset, it simply cannot discover the hardest topics that lie outside its search space.

Taken together, these results confirm that $\epsilon$-greedy is not only the most efficient algorithm (per \Cref{fig:03-algorithms_main}) but also the most \emph{accurate}: it converges quickly to faithful difficulty estimates by accumulating enough evidence per topic to distinguish genuine difficulty from variance-driven flukes, and it does so without artificially restricting the search space.
This analysis also complements Figure~2 in the main paper, which visualizes a related convergence phenomenon.

\begin{table}[h]
\centering
\small
\setlength{\tabcolsep}{4.5pt}
\begin{tabular}{r rrr rrr rrr rrr}
\toprule
& \multicolumn{3}{c}{\textbf{$\epsilon$-greedy ($\epsilon$=0.7)}} & \multicolumn{3}{c}{\textbf{Top-$D$ ($D$=10\%)}} & \multicolumn{3}{c}{\textbf{Brute-Force}} & \multicolumn{3}{c}{\textbf{Greedy (no cap)}} \\
\cmidrule(lr){2-4} \cmidrule(lr){5-7} \cmidrule(lr){8-10} \cmidrule(lr){11-13}
\textbf{Steps} & Est. & Act. & $\Delta$ & Est. & Act. & $\Delta$ & Est. & Act. & $\Delta$ & Est. & Act. & $\Delta$ \\
\midrule
1{,}000 & 16.91 & 13.26 & 3.65 & 31.78 & 11.79 & 19.99 & 32.93 & 12.33 & 20.60 & 32.93 & 12.33 & 20.60 \\
2{,}000 & 19.49 & 17.15 & 2.34 & 17.64 & 17.55 &  0.09 & 37.22 & 13.59 & 23.63 & 37.22 & 13.59 & 23.63 \\
3{,}000 & 21.62 & 19.66 & 1.96 & 17.98 & 17.95 &  0.03 & 40.07 & 15.07 & 25.00 & 40.07 & 15.07 & 25.00 \\
4{,}000 & 23.65 & 21.50 & 2.15 & 18.13 & 18.12 &  0.01 & 38.90 & 15.91 & 22.99 & 22.56 & 22.06 &  0.50 \\
5{,}000 & 23.78 & 23.66 & 0.12 & 18.22 & 18.19 &  0.03 & 37.33 & 16.27 & 21.06 & 22.66 & 22.65 &  0.01 \\
6{,}000 & 23.98 & 23.98 & 0.00 & 18.26 & 18.26 &  0.00 & 35.81 & 16.62 & 19.19 & 22.66 & 22.65 &  0.01 \\
\bottomrule
\end{tabular}
\caption{Estimated difficulty (search proxy) vs.\ actual difficulty (independently measured per-domain score) for all algorithms at six budget checkpoints. $\Delta = |\text{Estimated} - \text{Actual}|$ measures the gap; lower is better. \textbf{$\epsilon$-greedy} converges to near-zero $\Delta$ the fastest among all bandit methods. \textbf{Brute-force} never converges. \textbf{Top-$D$} converges early but plateaus at lower actual difficulty due to its restricted search space. \textbf{Greedy (no cap)} overshoots initially before correcting late in the search.}
\label{tab:estimated_vs_actual}
\end{table}

\section{Difficult Texts Rank Models Better}
\label{sec:ranking_efficiency}

Having a difficult testset at hand is beneficial for many reasons.
One of them is the higher discriminability; i.e. the ability of this data in distinguishing good from bad models and ranking them.
\citet{zouhar-etal-2025-select} find that difficult examples and examples where models have varying scores (high variance) tend to be better at efficiently ranking the models.
In this section, we confirm this for our method of obtaining difficult data.

\begin{figure}[t]
    \centering
    \includegraphics[width=1\linewidth]{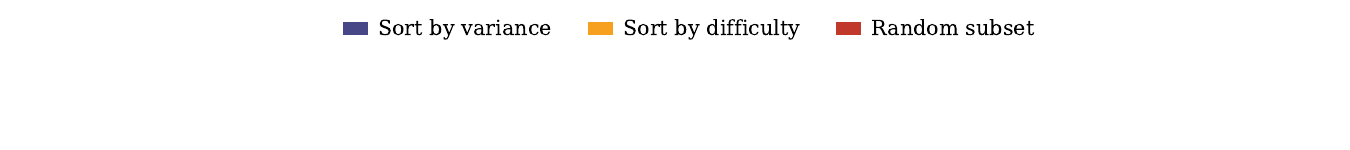}
    
    \vspace{-10mm}
    \includegraphics[width=1\linewidth]{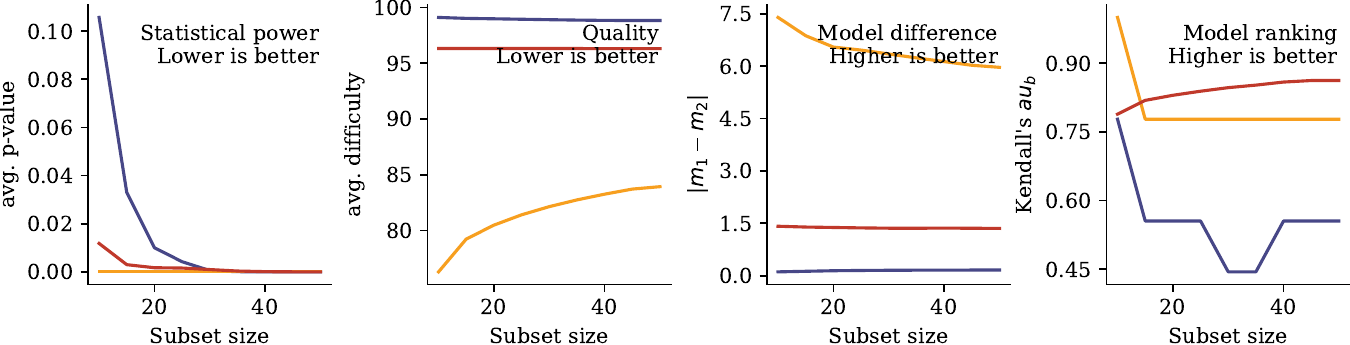}
    \caption{Various measures of subset utility: statistical power (p-value between adjacent models), average difficulty, average difference between adjacent models, ranking similarity between a subset and the whole testset (gold).
    }
    \label{fig:08-ranking_efficiency}
\end{figure}

In \Cref{fig:08-ranking_efficiency} we measure 4 key properties of a challenge set: (1) statistical discriminability, (2) average difficulty, (3) differences in average model scores, and (4) ranking on the subset with respect to the ranking on the whole set (proxy for gold model ranking).
Systematically, the difficult challenge set, based on our $\epsilon$-greedy selection strategy outperforms random test example selection.
We also include a challenge with the highest variance between models, as suggested by \citet{zouhar-etal-2025-select}, though this does not perform consistently well across all criteria.

\section{Other Search Algorithms}
\label{sec:other_algorithms}

\subsection{Contextual Bandit}
\begin{wrapfigure}{R}{0.36\textwidth}
\vspace{-2cm}

\raggedleft
\begin{center}
\includegraphics[width=0.9\linewidth]{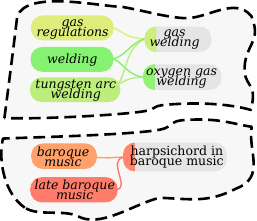}
\end{center}
\captionof{figure}{Illustration of neighbour effect on the selection of topics to sample from. Even though \textit{gas welding} and \textit{oxygen gas welding} topics were never sampled from, the neighbours suggest that those topics will not be difficult. In contrast, \textit{harpsichord in baroque music} will likely be difficult.}
\label{fig:contextual_illustration}

\vspace{-15mm}
\end{wrapfigure}
Contextual bandit makes an observation that topics that are similar to each other are likely going to have similar sample difficulty.
Therefore, even if a topic has never been sampled from, but the difficulty of all its neighbours is low, we do not have to spend budget on sampling from it.
This is shown in \Cref{fig:contextual_illustration}: The topic with difficult neighbours is prioritized over the topics with easy neighbours which are never going to be sampled from.
This extension of the greedy approach is known as contextual bandit where we can consider some features of the arms (in our case topics) for the selection.
Practically, we first make sure that all topics are scoreable (i.e. they have been sampled from or have at least two neighbours with samples).
Then, we interpolate between the topic's score and the scores of its neighbours based on the similarity.
For computing the similarity, we use the overlap in keywords (Jaccard index).

\begin{table}
\newcommand{\nansymbol}{\ensuremath{\bigcirc\hspace{-2.5mm}\text{\raisebox{-0.05mm}{\small?}}\hspace{2.5mm}}}

\hrule
\vspace{1mm}
\begin{algorithmic}[1]
\Statex \hspace{-5mm} \textbf{ContextualChoose}($T$: topics, $c$: cap):
\State \textbf{if} $\exists t \in T: \mathrm{ContextualScore}(t) = \nansymbol$
\State \quad \Return $\mathrm{uniform}(\{ t | t \in T, |t^\star| = 0\})$
\State \textbf{else}
\State \quad \Return $\arg \max_{t \in T, |t^\star|<c}  \mathrm{ContextualScore}(t)$
\end{algorithmic}
\vskip3pt
\begin{algorithmic}[1]
\Statex \hspace{-5mm} \textbf{ContextualScore}($t$: topic):
\State $N \gets \{ t_o | t_o \in T \setminus \{t\}, \mathrm{sim}(t, t_o) > 0  \,\wedge\, |t_o^\star| > 0 \}$
\hfill \graycomment{Find all scoreable neighbours}
\State \textbf{if} $|t^\star| = 0 \,\wedge\, |N| < 2$:
\hfill \graycomment{If this topic can not be scored, return \nansymbol}
\State \quad \Return \nansymbol
\State \textbf{else}
\State \quad $\beta = \begin{cases}
    0    & \text{if } |t^\star| = 0 \\
    0.5 & \text{if } |t^\star| = 1 \\
    1    & \text{if } |t^\star| \geq 2
\end{cases}$
\hfill \graycomment{Interpolate between own score and score by neighbours}
\State \quad \Return  $\beta \cdot \frac{\sum_{x \in t^\star} d_x}{|t^\star|}+ (1-\beta) \cdot \mathrm{softmax}(\langle \mathrm{sim}(t, t_o) | t_o \in N\rangle) \cdot \langle \frac{\sum_{x\in t_o} d_x}{|t_o^\star|} | t_o \in N\rangle$
\end{algorithmic}
\hrule
\vspace{2mm}
\captionof{algorithm}{Contextual bandit algorithm first makes all seed topics scoreable and then exploits the most difficult one limited by cap $c$.}
\label{alg:contextual_bandit}
\end{table}

\subsection{Upper Confidence Bound Bandit}

While there are many algorithms for multi-armed bandit, the biggest obstacle in our case in \Cref{fig:03-algorithms_main} is the cold-start and initialization phase needed (before all topics have been exploited at least once) for the greedy search.
Improvements that take into account the confidence, such as upper confidence bound bandit \citep{auer2002finite}, would still require sampling of all initial topics.
Their advantage would only be faster descent after the initialization phase, which is already steep for our method in comparison to the cost of the initialization phase.

\begin{table}[h]
\centering
\small
\begin{tabular}{lcc}
\toprule
\textbf{Error Category} & \textbf{Gemma-3-27b} & \textbf{Gemini-2.5-Pro} \\
\midrule
Fluency/Style & 14.2\% & 8.7\% \\
Terminology & 48.2\% & 37.6\% \\
Accuracy & 37.6\% & 53.7\% \\
\bottomrule
\end{tabular}
\caption{Error distribution on the same English$\rightarrow$Ukrainian test set for two models. The substantial variation confirms that the error profile is determined by each model's weaknesses, not by properties of our input.}
\label{tab:error_distribution_models}
\end{table}

\section{Per-Model Error Profile Analysis}
\label{sec:error_profile_analysis}

In \Cref{sec:error_analysis} of the main paper, we showed that the overall error distribution of our challenge benchmark is dominated by terminology (47.7\%) and accuracy (42.3\%) errors.
Here we break this down by model to demonstrate that the error profile is a property of each model's weaknesses rather than an artifact of the input data.

\Cref{tab:error_distribution_models} compares the error distributions of Gemma-3-27b and Gemini-2.5-Pro on the same English$\rightarrow$Ukrainian test set.
The two models exhibit markedly different error profiles despite being evaluated on identical inputs:

\paragraph{Gemma-3-27b exhibits higher fluency/style errors.}
Gemma-3-27b produces 14.2\% fluency/style errors compared to only 8.7\% for Gemini-2.5-Pro.
This suggests that the smaller, open-source model struggles more with generating fluent target-language text, particularly in morphologically rich languages like Ukrainian.
These errors manifest as grammatical mistakes, awkward phrasing, and register mismatches.

\paragraph{Gemini-2.5-Pro shifts toward accuracy errors.}
Gemini-2.5-Pro produces 53.7\% accuracy errors compared to 37.6\% for Gemma-3-27b.
While Gemini generates more fluent translations overall, it is more prone to meaning-altering errors: mistranslations, omissions, and additions that change the factual content.
This is consistent with the observation that frontier LLMs optimize for fluency at the potential cost of faithfulness \citep{kocmi-etal-2025-findings}.

\paragraph{Terminology errors are dominant for both models.}
Both models show high rates of terminology errors (48.2\% for Gemma, 37.6\% for Gemini), confirming that domain-specific terminology remains a shared weakness across model scales.
However, the relative balance shifts: Gemma's errors are more terminology-heavy, while Gemini's errors skew toward accuracy, with accuracy errors (53.7\%) substantially exceeding terminology errors (37.6\%).

These contrasting profiles confirm that the challenge examples discovered by our pipeline are structurally rich enough to trigger distinct failure modes in different systems.
If the inputs lacked complexity, all models would produce near-identical error distributions.
The observed variation validates that our benchmark probes genuine, model-specific weaknesses rather than superficial difficulty.

\section{Cost Analysis}
\label{sec:cost_full}

The cost breakdown in \Cref{tab:real_cost} (main paper) details the monetary cost at different scales.
For data generation, the average prompt length was 146 input tokens, resulting in an average generated source output of 94 tokens. Additional grounded search costs amounted to \$35 per 1,000 requests.
To collect 25 source sentences per topic, the Google Search-grounded Gemini model was prompted to extract relevant sentences from its search snippets, typically requiring an average of three queries per topic. For translation, the average prompt and output lengths were 107 and 76 tokens, respectively.
Quality estimation prompts and outputs averaged 310 and 595 tokens, respectively.

\section{Prompts}
\label{sec:prompts}

\newcommand{\promptexample}[2]{
\noindent \textbf{#1.}
{\fontsize{8.15pt}{8pt}\selectfont\tt #2 }
\medskip
}

\subsection{Topic Generation Prompt}
\label{sec:prompt_topic}

\promptexample{Topic generation}{
You are an expert ontologist specializing in {domain\_name}. Your task is to generate a comprehensive concept tree for this domain. Please adhere to the following specifications: \\
Output Format: Generate a single Python code block containing a nested dictionary representing the concept tree. \\\vskip1em
Tree Structure: \\
Root Node: The root of the tree must be '{domain\_name}'.\\
Depth: The tree must have a depth of {number\_of\_levels}, meaning there are {number\_of\_levels} levels of subtopics beneath the root node. \\
Branching Factor: Each parent node (non-leaf node) must generate exactly {branching\_factor} unique child nodes (subtopics). \\
Node Naming (Crucial): \\
Self-Contained: Each node name (the dictionary key) must be a self-contained and specific phrase suitable for a direct Google search. It must be fully understandable without knowing its parent topic. \\
Language: All node names must be in {language}. \\
Example Structure: The final output should follow this nested dictionary format (with no additional comments or text):\\
{ \\
\
"[Root Node Name]": \{ \\
\null\quad\quad"[Subtopic 1.1]": \{ \\
\null\quad\quad\quad"[Subtopic 1.1.1]": \{ \\
\null\quad\quad\quad\quad\# ... continue for specified depth \\
\null\quad\quad\quad\}, \\
\null\quad\quad\quad"[Subtopic 1.1.2]": \{ ... \}, \\
\null\quad\quad\quad"[Subtopic 1.1.3]": \{ ... \}, \\
\null\quad\quad\quad"[Subtopic 1.1.4]": \{ ... \} \\
\null\quad\quad\}, \\
\null\quad\quad"[Subtopic 1.2]": \{ ... \}, \\
\null\quad\quad"[Subtopic 1.3]": \{ ... \}, \\
\null\quad\quad"[Subtopic 1.4]": \{ ... \} \\
\null\quad} \\
\}
}

\subsection{Source Text Sampling Prompt}
\label{sec:prompt_sampling}

\promptexample{Sampling source text from topic}{
Please use Google search to find all relevant topics about {SEARCH\_KEY\_WORDS} in {LANG}. Then extract all relevant snippet contents in the format of JSON. Each extracted content should be approximately 20-40 words and distinct from each other. Please make sure to extract all relevant contents.\\
\{ \\
"extracted\_snippets": [ \\
\null\quad  \{ \\
\null\quad\quad"text": "content 1", \\
\null\quad\quad"source\_url": "http://example.com/source\_1" \\
\null\quad  \}, \\
\null\quad  ... \\
\null\quad  \{ \\
\null\quad\quad"text": "content n", \\
\null\quad\quad"source\_url": "http://example.com/source\_n" \\
\null\quad  \}, \\
\null\quad] \\
\}
}

\subsection{Quality Estimation Prompt}
\label{sec:prompt_qe}

\promptexample{Quality Estimation}{
Evaluate the quality of the translation on a scale from 0 to 100. Roughly: \\
100 - Perfect \\
95 - Excellent (closely aligned with the source) \\
80 - Very good (minor style choice) \\
60 - Fair (some inaccuracies or fluency errors) \\
40 - Poor (multiple inaccuracies or fluency errors) \\
0 - Inadequate (unrelated, completely wrong) \\
First, think about all the errors in the translation
and their severity (very briefly, max few words per error).
At the end, output a single line in the format like as follows:  \\
SCORE |||70.8||| \\
The last line is important because it will be matched with a regex, so make sure to use the |. \\
Don't think for too long (max 10 sentences). \\

SOURCE: |||{src}||| \\
TRANSLATION: |||{tgt}||| \\
}

\subsection{Translation Prompt}
\label{sec:prompt_translation}

\promptexample{Translation}{
You are a professional translator.
You are given a source text in {src\_lang}.
You need to translate the source text to {tgt\_lang}.
Don't include any other text except the translation.
Please output the translation between <START OF TRANSLATION> and </END OF TRANSLATION>. Source text: {src\_txt}
}

\subsection{KQA Question Sampling Prompt}
\label{sec:prompt_qa_sampling}

\promptexample{Sampling question from topic}{Task: Formulate a question in \{tar\_lang\} that requires a deep understanding of a given \{src\_lang\} paragraph.
Requirements:
* Context-Specific: The question must be answerable solely through information presented within the paragraph,
excluding general knowledge or common sense.
* Self-Contained: The question should be completely self-explanatory, providing all necessary context within
its phrasing. Assume the reader has no access to the paragraph when answering the question.
* Single Concrete Factual Detail:
- The question should not require multiple answers or involve listing multiple details.
- Avoid asking about opinions, interpretations.
- In you can't answer the question, prefer to generate another question.
- Focus on extracting a specific, concrete, factual detail that the paragraph directly states.
- Be specific:
- If you are asking about an entity be clear about it - Use full names for example.
- Mention expected granularity: If you are asking about a date, instead of asking "when", ask for a decade,
year, month, date etc. If you are asking about a location, instead of asking "where", ask for a country, state, city,
street, landmark etc.
- Avoid asking questions that their answers are acronyms.
- When formulating a question, you should assume that the person who answers this question does not have access to the paragraph. Avoid using phrases like "According to the paragraph".
- Make sure the answer is very short.
Even for non-English examples keep the convention of using the special words like "paragraph", "response",
"[BEGIN OF QUESTION]", "[END OF QUESTION]", "[BEGIN OF ANSWER]", "[END OF ANSWER]" for specifying the parts being generated.
Generate only the question and answer. No need to continue with additional examples.
Examples:
Paragraph: The Great Barrier Reef is the world's largest coral reef system, composed of over 2,900 individual
reefs and 900 islands stretching for over 2,300 kilometers (1,400 mi) over an area of approximately 344,400
square kilometers (133,000 sq mi). The reef is located in the Coral Sea, off the coast of Queensland, Australia.
The Great Barrier Reef can be seen from outer space and is the world's biggest single structure made by living
organisms.
Response:
[BEGIN OF QUESTION] Where is the Great Barrier Reef located?[END OF QUESTION]
[BEGIN OF ANSWER] Coral Sea, off the coast of Queensland, Australia[END OF ANSWER]
Paragraph: Die Cazoo Snookerweltmeisterschaft 2023 wurde vom 15. April bis 1. Mai im Crucible Theatre in
Sheffield ausgetragen. Mit ihr endete die Saison 2022/23 der World Snooker Tour.[1] Titelverteidiger Ronnie
O'Sullivan scheiterte im Viertelfinale gegen Luca Brecel. Der Belgier erreichte das Finale und schlug dort den
vierfachen Weltmeister Mark Selby mit 18:15. Brecel ist damit der erste Kontinentaleuropäer, der Weltmeister
wurde. In diesem Jahr wurden noch weitere Bestmarken in Bezug auf die 47-jährige „Crucible-Ära" aufgestellt.
Unter anderem übertraf Ronnie O'Sullivan mit seiner 31. Endrundenteilnahme die 30 Teilnahmen von Steve
Davis.[2] O'Sullivan erzielte auch sein 200. WM-Century-Break. Zweimal wurde ein Maximum Break erzielt,
was es 2008 bereits einmal gegeben hatte; das „perfekte Break" in einem WM-Finale gelang 2023 erstmals
Mark Selby.
Response:
[BEGIN OF QUESTION] Gegen wen verlor Ronnie O'Sullivan im Viertelfinale der Snooker-Weltmeisterschaft 2023?[END OF QUESTION]
[BEGIN OF ANSWER] Luca Brecel[END OF ANSWER]
Paragraph: {context}
Response:}

\subsection{KQA Answering Prompt}
\label{sec:prompt_qa_answer}

\promptexample{QA prompt}{You are an expert in knowledge question answering. Question: \{question\} Please answer the question between [BEGIN OF ANSWER] and [END OF ANSWER]. Make sure the answer is very short.}

\subsection{KQA Autorater Prompt}
\label{sec:prompt_qa_autorater}

\promptexample{QA Autorater Prompt}{You are an expert evaluator for Knowledge Question Answering. You will be provided with a Question, Context, Reference Answer, and a Model Response. Your goal is to judge the factual accuracy of the Model Response.

Evaluation Criteria:
Single Definitive Answer: If the question has only one correct answer, determine if the Model Response creates a semantic match with the Reference Answer.
Multiple Definitive Answers: If the question allows for various valid answers, assess factual correctness by verifying answer against the provided Context and the Reference Answer.

Answer 'yes' if the Model Response contains a factual mistake. Answer 'no' if the Model Response is factually correct. Answer 'yes' or 'no' only. At the end, output a single line in the format like as follows:
`ANSWER |||yes|||` or `ANSWER |||no|||`

QUESTION: |||{question}|||
CONTEXT: |||{context}|||
REFERENCE ANSWER: |||\{reference\_answer\}|||
MODEL RESPONSE: |||\{model\_response\}|||}

\subsection{AutoMQM Prompt}
\label{sec:prompt_automqm}

\promptexample{AutoMQM}{
You are an annotator for the quality of machine translation. Your task is to identify errors and assess the quality of the translation using MQM. Based on the source text (in \textless{}source\textgreater{}\textless{}/source\textgreater{} tags) and machine translation surrounded (in \textless{}translation\textgreater{}\textless{}/translation\textgreater{} tags), identify error types in the translation and classify them. The categories of errors are: accuracy (addition, mistranslation, omission, untranslated text, wrong language), fluency (character encoding, grammar, inconsistency, punctuation, register, spelling), style (awkward), terminology (inappropriate for context, inconsistent use), other. Each error, including omissions or untranslated content, is classified as one of three categories: critical, major, and minor. Critical errors inhibit comprehension of the text. Major errors disrupt the flow, but what the text is trying to say is still understandable. Minor errors are technically errors, but do not disrupt the flow or hinder comprehension. The source text must be fully covered and any omissions should also be annotated as errors. Please only include errors and no spans that do not contain errors.\\

You must conform to the following JSON schema:\\
\{\quad "type": "object",\\
\null\quad "properties": \{\\
\null\quad\quad   "errors": \{\\
\null\quad\quad\quad     "type": "array",\\
\null\quad\quad\quad     "items": \{\\
\null\quad\quad\quad\quad       "type": "object",\\
\null\quad\quad\quad\quad       "properties": \{\\
\null\quad\quad\quad\quad\quad         "error\_span": \{ "type": "string", "description": "The relevant input span where the error occurred." \},\\
\null\quad\quad\quad\quad\quad         "explanation": \{ "type": "string", "description": "A brief explanation of the error and its impact" \},\\
\null\quad\quad\quad\quad\quad         "error\_category": \{ "type": "string", "enum": ["accuracy", "fluency", "style", "terminology", "other"] \},\\
\null\quad\quad\quad\quad\quad         "error\_type": \{ "type": "string", "description": "The specific type of error within the category" \},\\
\null\quad\quad\quad\quad\quad         "severity": \{ "type": "string", "enum": ["critical", "major", "minor"] \}\\
\null\quad\quad\quad\quad       \},\\
\null\quad\quad\quad\quad       "required": ["explanation", "error\_category", "error\_type", "severity"]\\
\null\quad\quad\quad     \}\\
\null\quad\quad   \}\\
\null\quad \},\\
\null\quad "required": ["errors"]\\
\}\\

Please score the following input\\
\textless{}input\textgreater{}\\
\textless{}source\_language\textgreater{}\{src\_lang\}\textless{}/source\_language\textgreater{}\\
\textless{}source\textgreater{}\{source\}\textless{}/source\textgreater{}\\
\textless{}target\_language\textgreater{}\{tgt\_lang\}\textless{}/target\_language\textgreater{}\\
\textless{}translation\textgreater{}\{translation\}\textless{}/translation\textgreater{}\\
\textless{}/input\textgreater{}\\

Please respond in JSON without any introduction or explanation. Only the JSON response is required.\\
MQM JSON response:}

\begin{table}[h!]
\centering
\small
\resizebox{\linewidth}{!}{\parbox{\linewidth}{
\hspace{-1mm}\begin{tabular}{lp{2.2cm}@{\hspace{2mm}}rrp{7.97cm}}
\toprule
\multicolumn{2}{l}{\bf Existing domains} & \hspace{-10mm} \bf Difficulty$\bm{\uparrow}$ \hspace{-2mm} & \hspace{-3mm} \bf Words$\bm{\downarrow}$ \hspace{-2mm} &\bf Example \\
\midrule
\parbox[t]{4mm}{\multirow{4}{*}{\rotatebox[origin=c]{90}{\parbox{3.5cm}{\raggedleft WMT 2024\\ \citealp{kocmi-etal-2024-findings}}}}}
& Social
 & 10.1
 & 16
 & \smol In general I really like the interplay between the two games. The advantage of shortform stories is that you can "skip to the good part"...\\
& Literary
 & 8.3
 & 38
 & \smol The advancement of Humanity never ceased, even for a moment–during difficult times we grow and adapt once again. The cities are as prosperous as ever, and our technological advancement is rising. ...\\
& Speech
 & 8.3
 & 73
 & \smol Cheers, y'all. Now check it out. I really didn't even eat enough to be wiping my mouth, but I can tell you this, my mouth is salivating though. ....\\
& News
 & 6.4
 & 54
 & \smol "People Swimming in the Swimming Pool" from 2022 is one Vicente Siso artwork that will display at Tierra del Sol Gallery beginning Jan. 13. (photo courtesy of Vicente Siso)...\\\null\\
\parbox[t]{4mm}{\multirow{5}{*}{\rotatebox[origin=c]{90}{\parbox{3cm}{\raggedleft WMT 2025\\ \citealp{kocmi-etal-2025-findings}}}}}
& Speech
 & 17.3
 & 145
 & \smol Gotta watch a netflix show you feel me, but let me know down below. What show should i watch on netflix though? Because i'm i'm really having some trouble to find what show should ...\\
& News
 & 14.3
 & 95
 & \smol Some folks really do deserve a badge of honour for their pedantry (C8). Veronica Coyne of Springfield claims that "when bemoaning the loss of the express lane at Woolies "12 items or less,"...\\
& Social
 & 11.8
 & 98
 & \smol Another fine evening (ok not really, it's wet and drizzly, but) to continue exploring my stash of Rum from Japan 
Cor Cor again - this time the "Industrial" ....\\
& Literary
 & 9.9
 & 117
 & \smol It had been a remarkable twenty-year pro career, one that most players could only dream of. He wore a gleaming championship ring, a testament to his hard work and dedication ....\\
& Dialogue
 & 5.7
 & 179
 & \smol X: I am looking for a cheap hotel with free parking near Cambridge.
Y: I have multiple cheap hotels with free parking.  What part of town are you interested in staying in?...\\\null\\
\parbox[t]{4mm}{\vspace{-2mm}\multirow{5}{*}{\rotatebox[origin=c]{90}{\parbox{4cm}{\raggedleft FLORES-101\\ \citealp{nllbteam2022}}}}}
& Wikinews/disasters and accidents
 & 4.7
 & 18
 & \smol At 1:15 a.m. Saturday, according to witnesses, the bus was going through a green light when the car made a turn in front of it....\\
& Wikivoyage/travel
 & 4.6
 & 21
 & \smol Cold weather is perhaps the only real danger the unprepared will face....\\
& Wikinews/politics
 & 3.8
 & 21
 & \smol Mr Costello said that when nuclear power generation becomes economically viable, Australia should pursue its use....\\
& Wikinews/sports
 & 3.6
 & 18
 & \smol Mr Reid managed to drive the New Zealand's A1GP car, Black Beauty at speeds over 160km/h seven times over the bridge....\\
\bottomrule
\end{tabular}

\medskip

\begin{tabular}{p{5cm}@{\hspace{2mm}}rrp{6.1cm}}
\toprule
\bf Our topics & \hspace{-10mm} \bf Difficulty$\bm{\uparrow}$ \hspace{-2mm} & \hspace{-4mm} \bf Words$\bm{\downarrow}$ \hspace{-2mm} & \bf Example\\
\midrule
Incarceration Prison vs Jail
 & 39.5
 & 29
 & \smol Jails are short-term facilities for temporary detention, ... Prisons are long-term facilities for extended incarceration....\\
Leasehold Estates Tenancy for Years Periodic Tenancy
 & 29.6
 & 32
 & \smol Periodic Tenancy: A non-freehold estate that lasts only from period to period without having any definite duration that is longer than one ....\\
Future Interests Reversions Remainders Executory Interests
 & 29.5
 & 34
 & \smol Future interests are legal rights to property ownership that may become possessory later. They arise when a grantor conveys....\\
Removal Jurisdiction State to Federal Court
 & 21.2
 & 30
 & \smol For removal based on diversity jurisdiction, the amount in controversy must exceed \$75,000, and there must be complete diversity of ....\\
Victim Impact Statements Role
 & 21.0
 & 34
 & \smol Victim impact statements detail the emotional, physical, and financial consequences of a crime. They can be written or oral and are...\\
\bottomrule
\end{tabular}
}}
\caption{
Full comparison of topics from existing benchmarks and topics found by our $\epsilon$-greedy algorithm, with example texts.
All discovered topics are more difficult than existing benchmark subsets despite having lower average word counts (difficulty scales with length).
We include detailed case study of challenging source texts and model mistakes in \Cref{tab:example_ende,tab:example_encs,tab:example_enzh}.
}
\label{tab:04-topic_overview_full}
\end{table}

\clearpage

\begin{table}[h!]
\centering
\small
\resizebox{\linewidth}{!}{\parbox{\linewidth}{
\hspace{-1mm}\begin{tabular}{lp{2.2cm}@{\hspace{2mm}}rrp{7.97cm}}
\toprule
\multicolumn{2}{l}{\bf Existing domains} & \hspace{-10mm} \bf Difficulty$\bm{\uparrow}$ \hspace{-2mm} & \hspace{-3mm} \bf Words$\bm{\downarrow}$ \hspace{-2mm} &\bf Example \\
\midrule
\parbox[t]{4mm}{\multirow{4}{*}{\rotatebox[origin=c]{90}{\parbox{3.5cm}{\raggedleft WMT 2024\\ \citealp{kocmi-etal-2024-findings}}}}}
& News
 & 14.1
 & 58
 & \smol "People Swimming in the Swimming Pool" from 2022 is one Vicente Siso artwork that will display at Tierra del Sol Gallery beginning Jan. 13. (photo courtesy of Vicente Siso)...\\
 & Speech
 & 16.1
 & 80
 & \smol Cheers, y'all. Now check it out. I really didn't even eat enough to be wiping my mouth, but I can tell you this, my mouth is salivating though. ....\\
 & Literary
 & 19.0
 & 80
 & \smol The advancement of Humanity never ceased, even for a moment–during difficult times we grow and adapt once again. The cities are as prosperous as ever, and our technological advancement is rising. ...\\
 & Social
 & 39.9
 & 22
 & \smol In general I really like the interplay between the two games. The advantage of shortform stories is that you can "skip to the good part"...\\\null\\
\parbox[t]{4mm}{\multirow{5}{*}{\rotatebox[origin=c]{90}{\parbox{3cm}{\raggedleft WMT 2025\\ \citealp{kocmi-etal-2025-findings}}}}}
& Dialogue
 & 9.08
 & 191
 & \smol X: I am looking for a cheap hotel with free parking near Cambridge.
Y: I have multiple cheap hotels with free parking.  What part of town are you interested in staying in?...\\
& Literary
 & 14.4
 & 121
 & \smol It had been a remarkable twenty-year pro career, one that most players could only dream of. He wore a gleaming championship ring, a testament to his hard work and dedication ....\\
 & Social
 & 18.9
 & 76
 & \smol Another fine evening (ok not really, it's wet and drizzly, but) to continue exploring my stash of Rum from Japan 
Cor Cor again - this time the "Industrial" ....\\
& News
 & 23.9
 & 94
 & \smol Some folks really do deserve a badge of honour for their pedantry (C8). Veronica Coyne of Springfield claims that "when bemoaning the loss of the express lane at Woolies "12 items or less,"...\\
 & Speech
 & 25.6
 & 142
 & \smol Gotta watch a netflix show you feel me, but let me know down below. What show should i watch on netflix though? Because i'm i'm really having some trouble to find what show should ...\\\null\\
\parbox[t]{4mm}{\vspace{-2mm}\multirow{5}{*}{\rotatebox[origin=c]{90}{\parbox{4cm}{\raggedleft FLORES-101\\ \citealp{nllbteam2022}}}}}

& Wikinews/politics
 & 3.9
 & 22
 & \smol Mr Costello said that when nuclear power generation becomes economically viable, Australia should pursue its use....\\
& Wikinews/sports
 & 4.2
 & 19
 & \smol Mr Reid managed to drive the New Zealand's A1GP car, Black Beauty at speeds over 160km/h seven times over the bridge....\\
& Wikinews/disasters and accidents
 & 4.9
 & 18
 & \smol At 1:15 a.m. Saturday, according to witnesses, the bus was going through a green light when the car made a turn in front of it....\\
 & Wikivoyage/travel
 & 7.4
 & 21
 & \smol Cold weather is perhaps the only real danger the unprepared will face....\\
\bottomrule
\end{tabular}

\medskip

\begin{tabular}{p{5cm}@{\hspace{2mm}}rrp{6.1cm}}
\toprule
\bf Our topics & \hspace{-10mm} \bf Difficulty$\bm{\uparrow}$ \hspace{-2mm} & \hspace{-4mm} \bf Words$\bm{\downarrow}$ \hspace{-2mm} & \bf Example\\
\midrule
Incarceration Prison vs Jail
 & 39.5
 & 29
 & \smol Jails are short-term facilities for temporary detention, ... Prisons are long-term facilities for extended incarceration....\\
Leasehold Estates Tenancy for Years Periodic Tenancy
 & 29.6
 & 32
 & \smol Periodic Tenancy: A non-freehold estate that lasts only from period to period without having any definite duration that is longer than one ....\\
Future Interests Reversions Remainders Executory Interests
 & 29.5
 & 34
 & \smol Future interests are legal rights to property ownership that may become possessory later. They arise when a grantor conveys....\\
Removal Jurisdiction State to Federal Court
 & 21.2
 & 30
 & \smol For removal based on diversity jurisdiction, the amount in controversy must exceed \$75,000, and there must be complete diversity of ....\\
Victim Impact Statements Role
 & 21.0
 & 34
 & \smol Victim impact statements detail the emotional, physical, and financial consequences of a crime. They can be written or oral and are...\\
\bottomrule
\end{tabular}
}}
\caption{
From each domain in an existing benchmark, we selected the 25 most challenging sentences for three models: Google Translate, Gemma 3-27B, and Gemini 2.5 Pro.
Results show that our algorithm's most challenging topic, ``Incarceration: Prison vs. Jail,'' is comparable in difficulty to the most challenging subsets found across 12 domains in three widely used benchmarks.
Generally, all five of the most challenging topics discovered by our algorithm are more difficult than most subsets of the three benchmarks.
}
\label{tab:top_25-topic_overview}
\end{table}

\newcommand{\errorsrc}[1]{\hlc[blue!20]{#1}}
\newcommand{\error}[1]{\hlc[pink!50]{#1}}
\newcommand{\errorzh}[1]{\textcolor{pink!75!black!90!red}{#1}}

\clearpage
\begin{table}
\centering
\small
\begin{tabularx}{0.8\linewidth}{@{}p{3mm}X}
\toprule
\parbox[t]{2mm}{\multirow{5}{*}{\rotatebox[origin=c]{90}{\shortstack{English${\rightarrow}$Czech\\(Google Translate)}}}}
& \textbf{Topic:} Incarceration Prison vs Jail \\
& \textbf{Text:}
The key difference lies in the length of stay and jurisdiction. \errorsrc{Jails} are for temporary detention and operated locally, while \errorsrc{prisons} are for extended incarceration and managed by state or federal agencies. \\
& \textbf{Translation:}
Klíčový rozdíl spočívá v délce pobytu a jurisdikci. \error{Věznice} slouží k dočasnému zadržení a jsou provozovány na místní úrovni, zatímco \error{věznice} jsou určeny pro dlouhodobé věznění a spravují je státní nebo federální agentury. \\
& \textbf{Quality estimation:} 40{\scriptsize/100} \\
& \textbf{Error analysis:} The terms \textit{jails} and \textit{prisons} are both translated as \textit{věznice} which is a critical error when the point of the text is to explain the differences. The correct translation for \textit{jail} would be \textit{vazba}. \\\null\\[1em]
\parbox[t]{2mm}{\multirow{5}{*}{\rotatebox[origin=c]{90}{\shortstack{English${\rightarrow}$Czech\\(Gemini 2.5 pro)}}}}
& \textbf{Topic:} Preemption Doctrine Express vs Implied \\
& \textbf{Text:} Congress sometimes expressly provides that state laws on a given topic are \errorsrc{preempted} (this is known as “express \errorsrc{preemption}”). Other times state laws are held \errorsrc{preempted} by courts because it is impossible to comply with both a federal and state law or because the state law frustrates accomplishment of the federal law's objectives (this is known as “implied \errorsrc{preemption}”). \\
& \textbf{Translation:} Kongres někdy výslovně stanoví, že státní zákony týkající se daného tématu jsou \error{nadřazené} (toto je známé jako „výslovná \error{nadřazenost}“). Jindy jsou státní zákony považovány za \error{nadřazené} soudy, protože není možné dodržovat jak federální, tak státní zákon, nebo protože státní zákon maří dosažení cílů federálního zákona (toto je známé jako „implicitní \error{nadřazenost}“). \\
& \textbf{Quality estimation:} 15{\scriptsize/100} \\
& \textbf{Error analysis:} The term \textit{preempted} (meaning \textit{overriden}) is translated as \textit{nadřazené} (meaning \textit{superior}). This flips the meaning from the federal laws having priority to state laws having priority. The correct term is \textit{přednost}. \\\null\\[1em]
\parbox[t]{2mm}{\multirow{5}{*}{\rotatebox[origin=c]{90}{\shortstack{English${\rightarrow}$Czech\\(Gemma 3)}}}}
& \textbf{Topic:} Hearsay Evidence Admissibility Weight \\
& \textbf{Text:} \errorsrc{Hearsay} is an \errorsrc{out-of-court} statement offered in court to prove the truth of the matter asserted. It is generally inadmissible unless it falls under an exception. The weight of \errorsrc{hearsay} evidence is determined by the trier of fact. \\
& \textbf{Translation:} \error{Doslovný} výrok je \error{mimožaludní} prohlášení předložené u soudu k prokázání pravdivosti tvrzeného faktu. Obecně je nepřijatelný, pokud nespadá pod výjimku. Váha důkazů založených na \error{doslovném} výroku je určena soudem nebo porotou. \\
& \textbf{Quality estimation:} 25{\scriptsize/100} \\
& \textbf{Error analysis:} The term \textit{hearsay} (meaning that the evidence can not be established) is translated as \textit{doslovný} (meaning \textit{literal}), which flips the meaning. The word \textit{out-of-court} is translated as \textit{mimožáludní} which is not a word in Czech. \\
\bottomrule
\end{tabularx}
\caption{Example texts (inputs) and translations from English${\rightarrow}$Czech based on found difficult topics. Only major errors are discussed.}
\label{tab:example_encs}
\end{table}

\clearpage
\begin{table}
\centering
\small
\begin{tabularx}{0.8\linewidth}{@{}p{3mm}X}
\toprule
\parbox[t]{2mm}{\multirow{5}{*}{\rotatebox[origin=c]{90}{\shortstack{English${\rightarrow}$German\\(Gemini 2.5 pro)}}}}
& \textbf{Topic:} Removal Jurisdiction State to Federal Court \\
& \textbf{Text:}
\errorsrc{Removal} is the procedure that allows a defendant sued in state court to transfer the case to federal court if the plaintiff could have originally filed the case in federal court. \\
& \textbf{Translation:}
Unter \error{Abschiebung} versteht man das Verfahren, das es einem vor einem Staatsgericht verklagten Beklagten ermöglicht, den Fall an ein Bundesgericht zu überweisen, wenn der Kläger den Fall ursprünglich vor einem Bundesgericht hätte einreichen können. \\
& \textbf{Quality estimation:} 25{\scriptsize/100} \\
& \textbf{Error analysis:} The term \textit{removal} is translated as \textit{Abschiebung} (meaning \textit{deportation}). The correct term is \textit{Überweisung}. \\\null\\[1em]
\parbox[t]{2mm}{\multirow{5}{*}{\rotatebox[origin=c]{90}{\shortstack{English${\rightarrow}$German\\(Google Translate)}}}}
& \textbf{Topic:} Incarceration Prison vs Jail \\
& \textbf{Text:}
\errorsrc{Jails} are typically locally operated, short-term facilities for those awaiting trial/sentencing or serving misdemeanor sentences (under a year). \errorsrc{Prisons} are state or federally operated long-term facilities.\\
& \textbf{Translation:}
\error{Gefängnisse} sind in der Regel lokal betriebene, kurzfristige Einrichtungen für Personen, die auf ihren Prozess/ihre Verurteilung warten oder eine Freiheitsstrafe für ein Vergehen (unter einem Jahr) verbüßen. \error{Haftanstalten} sind staatlich oder bundesstaatlich betriebene Langzeiteinrichtungen. \\
& \textbf{Quality estimation:} 40{\scriptsize/100} \\
& \textbf{Error analysis:} The terms \textit{jails} and \textit{prisons} are translated as \textit{Gefängnisse} and \textit{Haftanstalten}. While both of those terms could be jail and prison depending on the context, \textit{Gefängnisse} implies a longer-term facility while \textit{Haftanstalten} is used for shorter-term purposes. The intended meaning is ths reversed. \\\null\\[1em]
\parbox[t]{2mm}{\multirow{5}{*}{\rotatebox[origin=c]{90}{\shortstack{English${\rightarrow}$German\\(Gemma 3)}}}}
& \textbf{Topic:} Hearsay Evidence Admissibility Weight \\
& \textbf{Text:} \errorsrc{Hearsay} is defined as an out of court statement, made in court, to prove the truth of the matter asserted. In other words, hearsay is evidence of a statement. \\
& \textbf{Translation:}
\error{Klatsch} ist definiert als eine außergerichtliche Aussage, die vor Gericht gemacht wird, um die Richtigkeit des behaupteten Sachverhalts zu beweisen. Mit anderen Worten, Klatsch ist der Beweis einer Aussage... \\
& \textbf{Quality estimation:} 25{\scriptsize/100} \\
& \textbf{Error analysis:} The term \textit{hearsay} (meaning that the evidence can not be established) is translated as \textit{Klatsch} (meaning \textit{gossip}). While informally this would be acceptable, it is improper terminology in this judicial context. The correct term is \textit{Hörensagen}. \\
\bottomrule
\end{tabularx}
\caption{Example texts (inputs) and translations from English${\rightarrow}$German based on found difficult topics. Only major errors are discussed.}
\label{tab:example_ende}
\end{table}

\newcommand{\chinese}[1]{\begin{CJK*}{UTF8}{gbsn} #1 \end{CJK*}}
\clearpage
\begin{table}
\centering
\small
\begin{tabularx}{0.8\linewidth}{@{}p{3mm}X}
\toprule
\parbox[t]{2mm}{\multirow{5}{*}{\rotatebox[origin=c]{90}{\shortstack{English${\rightarrow}$Chinese\\(Gemini 2.5 pro)}}}}
& \textbf{Topic:} Preemption Doctrine Express vs Implied \\
& \textbf{Text:}
\errorsrc{Express preemption} occurs when Congress includes language in a federal statute explicitly stating that state law is preempted. An example is the Federal Aviation Administration Authorization Act of 1994.\\
& \textbf{Translation:}
\chinese{当国会在联邦法规中明确规定州法律被\errorzh{优先适用}时，就会发生明示\errorzh{优先适用}。例如 1994 年的《联邦航空管理局授权法》。}
\\
& \textbf{Quality estimation:} 45{\scriptsize/100} \\
& \textbf{Error analysis:} The term \textit{Express preemption} is translated as \chinese{优先适用} (meaning \textit{applied with priority}). The correct term is \chinese{明示取代}.\\\null\\[1em]
\parbox[t]{2mm}{\multirow{5}{*}{\rotatebox[origin=c]{90}{\shortstack{English${\rightarrow}$Chinese\\(Google Translate)}}}}
& \textbf{Topic:} Incarceration Prison vs Jail \\
& \textbf{Text:}
\errorsrc{Jails} are local facilities confining persons before or after adjudication. \errorsrc{Prisons} are state/federal facilities confining convicted felons. Jails have diverse populations; prisons hold sentenced felons.\\
& \textbf{Translation:}
\chinese{\errorzh{监狱}是地方性机构，用于关押判决前或判决后的人员。\errorzh{监狱}是州/联邦机构，用于关押已定罪的重罪犯。\errorzh{监狱}的关押对象多种多样；\errorzh{监狱}关押的是已判刑的重罪犯。} \\
& \textbf{Quality estimation:} 40{\scriptsize/100} \\
& \textbf{Error analysis:} The terms \textit{jails} and \textit{prisons} are both translated as \chinese{监狱}. While \chinese{监狱} can sometimes encompass both concepts in a very general sense or be used colloquially, in a legal/technical context like this, distinguishing them is crucial. ``\textit{Jails}'' are typically translated as \chinese{看守所}. ``\textit{Prisons}'' are correctly translated as \chinese{监狱}\\\null\\[1em]
\parbox[t]{2mm}{\multirow{5}{*}{\rotatebox[origin=c]{90}{\shortstack{English${\rightarrow}$Chinese\\(Gemma 3)}}}}
& \textbf{Topic:} Battery Harmful or Offensive Contact \\
& \textbf{Text:} \errorsrc{Battery} is the intentional act of causing harmful or offensive contact. Unlike assault (which can be just a threat), battery requires actual physical contact. This contact need not cause injury, only be offensive. \\
& \textbf{Translation:}
\chinese{\errorzh{电池}是指故意造成有害或冒犯性的接触行为。与袭击（仅可以是威胁）不同，\errorzh{电池}需要实际的身体接触。这种接触不一定需要造成伤害，只需具有冒犯性即可。}\\
& \textbf{Quality estimation:} 15{\scriptsize/100} \\
& \textbf{Error analysis:} Mistranslation of ``\textit{Battery}'' (legal term) as \chinese{电池} (electric battery): Critical/Major. At legal term, Battery means the completed act of unwanted physical contact. \\
\bottomrule
\end{tabularx}
\caption{Example texts (inputs) and translations from English${\rightarrow}$Chinese based on found difficult topics. Only major errors are discussed.}
\label{tab:example_enzh}
\end{table}


\clearpage

\section*{NeurIPS Paper Checklist}

\begin{enumerate}

\item {\bf Claims}
    \item[] Question: Do the main claims made in the abstract and introduction accurately reflect the paper's contributions and scope?
    \item[] Answer: \answerYes{}
    \item[] Justification: The abstract and introduction clearly state that we formalize challenge benchmark construction as a MAB problem and demonstrate it on two tasks (machine translation and knowledge QA), with specific quantitative claims (6\% search space, 100$\times$ cost reduction) supported by experiments.

\item {\bf Limitations}
    \item[] Question: Does the paper discuss the limitations of the work performed by the authors?
    \item[] Answer: \answerYes{}
    \item[] Justification: We address proxy metric robustness in Section~\ref{sec:cross_metric} and discuss cost trade-offs at different scales in Section~\ref{sec:cost}. Our framework is intentionally task-agnostic and we demonstrate it on two diverse tasks (machine translation and knowledge QA) as a proof of concept, with the conclusion noting that further applications are a natural direction.

\item {\bf Theory assumptions and proofs}
    \item[] Question: For each theoretical result, does the paper provide the full set of assumptions and a complete (and correct) proof?
    \item[] Answer: \answerNA{}
    \item[] Justification: The paper does not include formal theoretical results. Our contribution is empirical, demonstrating the effectiveness of MAB formulations for difficult example discovery.

    \item {\bf Experimental result reproducibility}
    \item[] Question: Does the paper fully disclose all the information needed to reproduce the main experimental results of the paper to the extent that it affects the main claims and/or conclusions of the paper (regardless of whether the code and data are provided or not)?
    \item[] Answer: \answerYes{}
    \item[] Justification: All search algorithm pseudocode is provided (Algorithms 1--4), prompts are detailed in Appendix~\ref{sec:prompts}, hyperparameters are specified ($\epsilon$=0.7), models are identified (Gemini 2.5 Pro, Gemma 3, Google Translate), and cost breakdowns are given. Crucially, we run the entire pipeline using the open-source model Gemma3-27B, ensuring the framework is fully reproducible by the open-source community without requiring access to proprietary APIs.

\item {\bf Open access to data and code}
    \item[] Question: Does the paper provide open access to the data and code, with sufficient instructions to faithfully reproduce the main experimental results, as described in supplemental material?
    \item[] Answer: \answerYes{}
    \item[] Justification: We plan to release the curated dataset of difficult topics and the search pipeline code upon publication. All prompts and algorithm pseudocode are provided in the paper.

\item {\bf Experimental setting/details}
    \item[] Question: Does the paper specify all the training and test details (e.g., data splits, hyperparameters, how they were chosen, type of optimizer) necessary to understand the results?
    \item[] Answer: \answerYes{}
    \item[] Justification: Section~\ref{sec:experiments} specifies the models, language pairs, topic set sizes, $\epsilon$ values, and budget constraints. Cost analysis is provided in Section~\ref{sec:cost} and Appendix~\ref{sec:cost_full}.

\item {\bf Experiment statistical significance}
    \item[] Question: Does the paper report error bars suitably and correctly defined or other appropriate information about the statistical significance of the experiments?
    \item[] Answer: \answerYes{}
    \item[] Justification: All experiments are conducted across 10 random seeds, and the main algorithm comparison figures report mean performance with variance across seeds at varying budgets. Cross-metric validation (Table~\ref{tab:cross_metric_guide}) demonstrates consistency across independent metrics. Oracle baselines provide upper bounds.

\item {\bf Experiments compute resources}
    \item[] Question: For each experiment, does the paper provide sufficient information on the computer resources (type of compute workers, memory, time of execution) needed to reproduce the experiments?
    \item[] Answer: \answerYes{}
    \item[] Justification: Section~\ref{sec:cost} provides detailed monetary cost breakdowns for API calls at various scales. Appendix~\ref{sec:cost_full} gives token-level cost analysis.
    
\item {\bf Code of ethics}
    \item[] Question: Does the research conducted in the paper conform, in every respect, with the NeurIPS Code of Ethics \url{https://neurips.cc/public/EthicsGuidelines}?
    \item[] Answer: \answerYes{}
    \item[] Justification: Our work focuses on benchmark evaluation methodology. The collected data comes from publicly available Internet sources and is used solely for evaluation purposes.

\item {\bf Broader impacts}
    \item[] Question: Does the paper discuss both potential positive societal impacts and negative societal impacts of the work performed?
    \item[] Answer: \answerYes{}
    \item[] Justification: Static benchmarks have saturated: they no longer effectively measure state-of-the-art models' performance or expose the tail of the performance distribution where models still fail. This work introduces a fully automatic framework that searches the Internet at scale to construct challenging benchmarks without human curation, reorienting evaluation from passive measurement on fixed corpora to active, adversarial search. By searching beyond saturation, we demonstrate that challenging examples can be identified at over 100$\times$ lower cost than exhaustive evaluation. Our framework is task-agnostic: we demonstrate it on both machine translation and knowledge question answering, and it extends naturally to any task where difficulty can be estimated. This establishes a sustainable path for benchmark construction that evolves alongside model capabilities, ensuring evaluation exposes genuine weaknesses rather than measuring memorization.
    
\item {\bf Safeguards}
    \item[] Question: Does the paper describe safeguards that have been put in place for responsible release of data or models that have a high risk for misuse (e.g., pre-trained language models, image generators, or scraped datasets)?
    \item[] Answer: \answerNA{}
    \item[] Justification: Although our contribution is primarily a benchmarking framework, we have applied safety and privacy filters to all collected data. Any data released publicly will be required to pass these filters, ensuring that unsafe or privacy-violating content is excluded from the released dataset.

\item {\bf Licenses for existing assets}
    \item[] Question: Are the creators or original owners of assets (e.g., code, data, models), used in the paper, properly credited and are the license and terms of use explicitly mentioned and properly respected?
    \item[] Answer: \answerYes{}
    \item[] Justification: All existing benchmarks (WMT, FLORES) and models (Gemini, Gemma, Google Translate) are properly cited. The ECLeKTic benchmark is cited for KQA experiments.

\item {\bf New assets}
    \item[] Question: Are new assets introduced in the paper well documented and is the documentation provided alongside the assets?
    \item[] Answer: \answerYes{}
    \item[] Justification: The curated challenge benchmark and search pipeline will be released with documentation. The data generation pipeline is fully described in the paper with prompts and pseudocode.

\item {\bf Crowdsourcing and research with human subjects}
    \item[] Question: For crowdsourcing experiments and research with human subjects, does the paper include the full text of instructions given to participants and screenshots, if applicable, as well as details about compensation (if any)? 
    \item[] Answer: \answerNA{}
    \item[] Justification: This paper does not involve crowdsourcing or research with human subjects.

\item {\bf Institutional review board (IRB) approvals or equivalent for research with human subjects}
    \item[] Question: Does the paper describe potential risks incurred by study participants, whether such risks were disclosed to the subjects, and whether Institutional Review Board (IRB) approvals (or an equivalent approval/review based on the requirements of your country or institution) were obtained?
    \item[] Answer: \answerNA{}
    \item[] Justification: This paper does not involve research with human subjects.

\item {\bf Declaration of LLM usage}
    \item[] Question: Does the paper describe the usage of LLMs if it is an important, original, or non-standard component of the core methods in this research? Note that if the LLM is used only for writing, editing, or formatting purposes and does \emph{not} impact the core methodology, scientific rigor, or originality of the research, declaration is not required.
    \item[] Answer: \answerYes{}
    \item[] Justification: LLMs (Gemini 2.5 Pro) are a core component of our pipeline: they are used for topic generation, text sampling via Google Search grounding, translation, and quality estimation. This is fully described in Section~\ref{sec:methods} and the prompts are provided in Appendix~\ref{sec:prompts}. We additionally demonstrate the entire pipeline using the open-source model Gemma3-27B (Section~\ref{sec:open_source}), showing that our framework is independent of the model backbone and accessible to the broader open-source community.

\end{enumerate}

\end{document}